\pdfoutput=1

\documentclass[11pt]{article}

\usepackage[final]{acl}

\usepackage{times}
\usepackage{latexsym}

\usepackage[T1]{fontenc}

\usepackage[utf8]{inputenc}

\usepackage{microtype}

\usepackage{inconsolata}

\usepackage{graphicx}

\usepackage{booktabs}
\usepackage{multirow}
\usepackage{amsmath}
\usepackage{amsfonts}
\usepackage{float} 
\usepackage{ulem}
\usepackage{algorithm} 
\usepackage{algorithmic} 
\usepackage{multicol}
\usepackage{enumitem}
\usepackage{amsthm} 
\newtheorem{theorem}{Theorem}
\newtheorem{lemma}{Lemma}
\newtheorem{insight}{Insight}
\newtheorem{definition}{Definition}
\usepackage{setspace}
\usepackage{ulem} 

\title{Multimodal Transformers are Hierarchical Modal-wise Heterogeneous Graphs}

\author{
 \textbf{Yijie Jin\textsuperscript{1}},
 \textbf{Junjie Peng\textsuperscript{1,\thanks{Corresponding author}}},
 \textbf{Xuanchao Lin\textsuperscript{1}},
 \textbf{Haochen Yuan\textsuperscript{1}},
 \textbf{Lan Wang\textsuperscript{1}},
 \textbf{Cangzhi Zheng\textsuperscript{1}},
\\
  \texttt{\{jyj2431567, jjie.peng, linxuanchao, yuanhc, Wanglan1997, cangzhizheng\}@shu.edu.cn} 
\\
 \textsuperscript{1}School of Computer Engineering and Science, Shanghai University
\\
}

\begin{document}
\maketitle
\begin{abstract}
Multimodal Sentiment Analysis (MSA) is a rapidly developing field that integrates multimodal information to recognize sentiments, and existing models have made significant progress in this area. The central challenge in MSA is multimodal fusion, which is predominantly addressed by Multimodal Transformers (MulTs). Although act as the paradigm, MulTs suffer from efficiency concerns. In this work, from the perspective of efficiency optimization, we propose and prove that MulTs are hierarchical modal-wise heterogeneous graphs (HMHGs), and we introduce the graph-structured representation pattern of MulTs. Based on this pattern, we propose an Interlaced Mask (IM) mechanism to design the Graph-Structured and Interlaced-Masked Multimodal Transformer (GsiT). It is formally equivalent to MulTs which achieves an efficient weight-sharing mechanism without information disorder through IM, enabling All-Modal-In-One fusion with only 1/3 of the parameters of pure MulTs. A Triton kernel called Decomposition is implemented to ensure avoiding additional computational overhead. Moreover, it achieves significantly higher performance than traditional MulTs. To further validate the effectiveness of GsiT itself and the HMHG concept, we integrate them into multiple state-of-the-art models and demonstrate notable performance improvements and parameter reduction on widely used MSA datasets.
\end{abstract}

\section{Introduction}

\begin{figure}[t]
\centering
  \includegraphics[width=\linewidth]{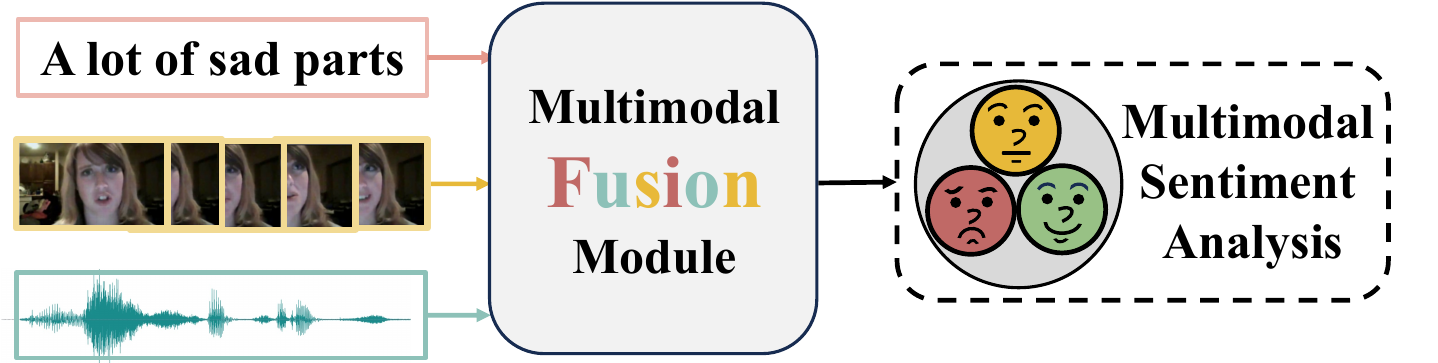}
  \caption{An example of multimodal sentiment analysis: end-to-end discriminative task pipeline.}
    \label{fig:example}
\end{figure}

With the growing ubiquity of diverse social media platforms such as YouTube and TikTok, users now express sentiments through various forms of information, including text, video, and audio. To achieve more natural human-computer interactions, multimodal sentiment analysis (MSA) has become a popular research area  \citep{if:survey}. MSA is briefly exemplified in Figure \ref{fig:example}.

\begin{figure*}[htpb]
\centering
  \includegraphics[width=0.9\linewidth]{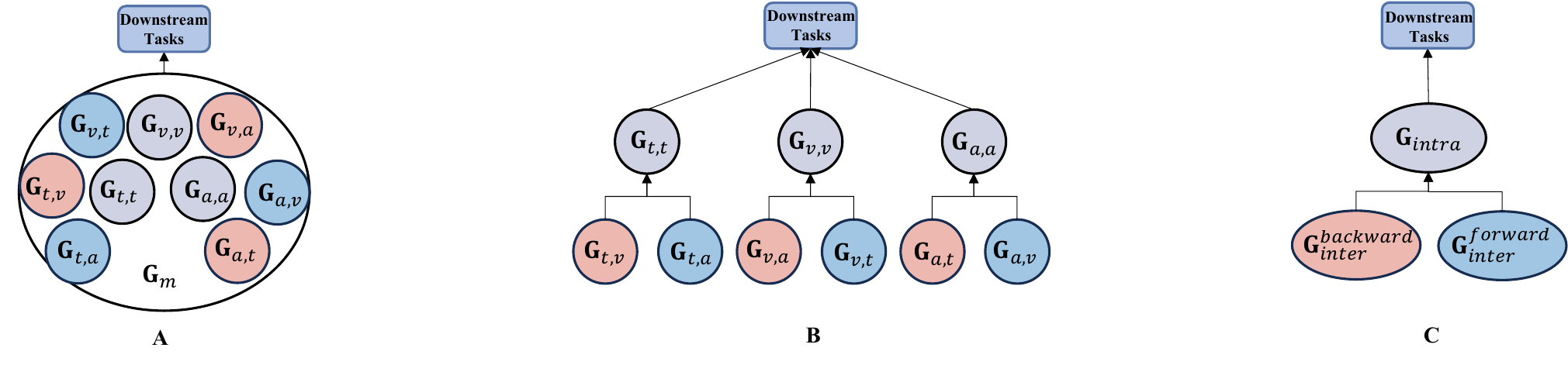}
  \caption{Graph structure comparison. A: Naive graph structure constructed by concatenated multimodal sequences. B: Forest structure of MulTs, constructed by decoupled bi-modality combinations. C: Tree structure of GsiT, constructed by concatenated multimodal sequences and IM machanism.}
\label{fig:graph_structure_comp}
\end{figure*}

The main challenge of MSA is to integrate heterogeneous data containing different sentiment information, thus achieving effective sentiment analysis. The practical manifestations of these challenges primarily lie in the performance of multimodal fusion methods, the representation capacity of multimodal features, and the robustness of the model. To address these challenges, methods of MSA involve designing effective multimodal fusion methods  \citep{emnlp:tfn, acl:mult, emnlp:almt, eswa:fnenet} to fully integrate heterogeneous data, and developing representation learning-based methods  \citep{aaai:self-mm, acl:confede, taffc:mtmd} to enhance unimodal information and model robustness. Among these, multimodal fusion is the core issue of MSA and also the focus of this paper. 

In the realm of multimodal fusion, Multimodal Transformer (MulT) \citep{acl:mult} and its enhanced successors \citep{aaai:hhmpn, emnlp:almt, mm:acformer, ipm:cmhfm, eswa:fnenet, naacl:mmml}, collectively known as MulTs, have shown significant effectiveness in MSA. Despite their status as the prevailing paradigm, the extensive use of Cross-Modal Attention (CMA) and Multi-Head Self-Attention (MHSA) mechanisms leads to inefficiencies in MulTs. Since MSA is an end-to-end discriminative task, it is imperative to reduce system overhead and improve model performance for the practical implementation of future MSA systems. Thus, the main objective of this work is to introduce a more efficient paradigm for MSA. Additional related works can be found in Appendix \ref{relatedworks}.

In this work, from the perspective of efficiency optimization, we discover and prove the theoretical equivalence between CMA/MHSA and Graph Attention Networks (GAT) \citep{iclr:gat}, where GAT uses multi-head attention as the aggregation function. Specifically, CMA is equivalent to a unidirectional complete bipartite graph of bi-modality combinations, while MHSA is equivalent to a directed complete graph of uni-modality. Based on this, MulTs can be defined as a forest composed of three independent trees. Each tree is constructed from three subgraphs, with hierarchical relationships constrained by a complex system of multiple functions. This mathematical representation formally defines the theorem that MulTs are hierarchical modal-wise heterogeneous graphs (HMHGs), as shown in Figure \ref{fig:graph_structure_comp}.

Based on the above theorem, we identify the redundancy in MulTs' model parameters and their potential for compression while preserving theoretical equivalence. Leveraging this discovery, we propose the Graph-Structured Interlaced-Masked Multimodal Transformer (GsiT) by compressing a forest composed of three independent trees into a single shared tree. GsiT introduces a novel Interlaced Mask (IM) mechanism for multimodal weight sharing, enabling All-Modal-In-One fusion without information disorder. Furthermore, we implement a Triton kernel named Decomposition to maintain efficiency. With only 1/3 of the parameters of traditional MulTs, GsiT maintains theoretical consistency with the MulTs' paradigm. Comprehensive experimental analysis reveals that GsiT outperforms traditional MulTs significantly in the same experimental setup, boasting a substantial edge in efficiency.

To validate the effectiveness and transferability of GsiT, we conducted comprehensive evaluations on the most widely used multimodal sentiment analysis datasets, including CMU-MOSI \citep{arxiv:mosi}, CMU-MOSEI \citep{acl:mosei}, CH-SIMS \citep{acl:ch-sims} (for multilingual adaptability), and MIntRec \citep{mm:mintrec} (for broader multimodal domains). Our findings show that GsiT not only outperforms as a backbone-level model but also that baseline models incorporating the HMHG concept achieve significant improvements in overall performance. 

\section{Insights}

MulTs facilitate multimodal fusion by breaking down multimodal data into pairs of modalities for processing. By creating various combinations of these bi-modality units, MulTs ensure a comprehensive integration of heterogeneous data.

This approach can be recognized as a hierarchical and graph-structured fusion method. To better illustrate, we first formally define hierarchical relationships.

\begin{definition}
    \label{definition1}
    Let $S$ and $T$ be two sets, belonging to the domain $\mathbb{X}$ and the range $\mathbb{Y}$, respectively, i.e., $S \subseteq \mathbb{X}$ and $T \subseteq \mathbb{Y}$. If there exists a mapping $f: \mathbb{X} \to \mathbb{Y}$ such that for any element $s_i \in S$, its corresponding mapped value $f(s_i) \in T$ depends on some subset $S_i \subseteq S$, and these dependency relationships can be constructed recursively, then the dependency relationship between $S$ and $T$ is called a \textbf{hierarchical relationship}. Furthermore, this hierarchical relationship can be represented by a \textbf{directed tree structure}, where vertices represent elements in the sets, and edges represent dependency relationships.
\end{definition}

To better define this type of model, we propose the following theorem:

\begin{theorem}
\label{theorem1}
Multimodal Transformers are hierarchical modal-wise heterogeneous graphs.
\end{theorem}

The formal theorem and its corresponding proof can be found in Section \ref{theorem}.

Since MSA task systems are end-to-end discriminative systems, we give the following insight. 

\begin{insight}
    \label{insight1}
    For MSA systems, the resource savings achieved by designing low-cost, high-performance models, which lead to overall performance improvements, are more significant in some aspects than the accuracy improvements brought by using large models with high representation capacity.
\end{insight}

\section{Multimodal Transformers as Graphs}

\label{theorem}

We first define modality text, vision, and audio as $t,v,a$, while multimodality as $m$. Assuming multimodal sequences $\mathcal{V}_{u_1} \in \mathbb{R}^{T_{u_1} \times d_{u_1}}$, where $u_1\in {\{m,t,v,a\}}$, $T_{u_1}$ denotes the temporal dimension (number of vertices), $d_{u_1}$ denotes feature dimension. Those sequences are then concatenated into a single sequence $\mathcal{V}_m = [\mathcal{V}_t; \mathcal{V}_v; \mathcal{V}_a]^{\top}$. $\mathcal{V}_m$ is treated as the multimodal graph embedding (MGE), which is also regarded as the multimodal vertex set. We define $\mathcal{W}_{u_2} \in \mathbb{R}^{d_{u_2} \times d_{u_2}^f}$, where $u_2 \in \{q, k, v\}$, $d_{u_2}$ is the original feature dimension of the vertices, $d_{u_2}^f$ is the attention feature dimension, as the projection weights for queries, keys, and values of $\mathcal{V}_m$. 

\subsection{Modal-wise Heterogeneous Graphs}

We first introduce a lemma as follows. For detailed proof, please refer to Appendix \ref{sec:Lemma and Proof}.

\begin{lemma}
\label{lemma1}
    The multi-head cross-modal attention mechanism is equivalent to the aggregation of unidirectional complete bipartite graphs of bi-modality combination; the multi-head self-attention mechanism is equivalent to the aggregation of directed complete graphs of uni-modalities.
\end{lemma}

Based on Lemma \ref{lemma1}, we decompose multi-head self-attention (MHSA) and multi-head cross-modal attention (CMA) into two steps of functions.

\vspace{-8pt}
\begin{equation*}
    \begin{aligned}
        &\textbf{Generate Adjacency Matrix:} & MHSA_1, CMA_1 \\
        &\textbf{Aggregation Operation:} & MHSA_2,  CMA_2 
    \end{aligned}
\end{equation*} 

The structure naive modal-wise heterogeneous graphs (MHGs) $\textbf{G}_m$ is defined as depicted in Figure \ref{fig:graph_structure_comp}. The attention map $\mathcal{G}$ is formulated as an adjacency matrix resulting from $MHSA_1$ and $CMA_1$, which effectively represent a set of edges with corresponding weights. Specifically, for $\mathcal{G}^{i,j}$, where $\{i,j\} \in \{t,v,a\}$: When $i \neq j$, it signifies the adjacency matrix of a complete bipartite graph of the bi-modality combination of $i$ and $j$, with the directionality being from $j$ to $i$; When $i = j$, it represents the adjacency matrix of the directed complete graph of uni-modality $i$; Specially, $\mathcal{G}^m$ denotes the adjacency matrix compose of all the $\mathcal{G}^{i,j}$.

Here, for features, we define $\mathcal{X}_m \in \mathbb{R}^{3d_m}$, where $d_m$ denotes the feature dimension, as the fusion output. In constructed MGE, we define $d_{\{t,v,a\}} = d_m$ to concatenate multimodal sequences. For functions, we define multi-layer perceptrons (also known as feed-forward networks) as a function $MLP$, function composition as $\circ$,  the final linear transformation as a function $f$, and $Split$ function, the concatenation operation on feature dimension as $\parallel$, which splits concatenated multimodal sequences into separated ones according to their original lengths.  

\vspace{-8pt}
\begin{equation}
    \begin{aligned}
        &\textbf{G}_m = (\mathcal{V}_m, \mathcal{G}^m), \quad \mathcal{G}^m = MHSA_1(\mathcal{V}_m) \\
        &a = MLP\circ MHSA_2 \\
        &\mathcal{X}_m = f(\parallel Split(a(\textbf{G}_m))[-1])
\end{aligned}
\end{equation}

\subsection{MulTs are Hierarchical MHGs}

In this section, we define the graph representation of MulTs. Assume the three indices follow the form $\quad i \in \{t, v, a\}$, $j \in \{t, v, a\} \setminus \{i\}$, $p \in \{t, v, a\} \setminus \{i, j\}$. Here, we define $H_u \in \mathbb{R}^{d_u}$, where $u\in {\{i,j,p\}}$ as the final state vector.

\vspace{-8pt}
\begin{equation}
    \begin{aligned}
        \label{g_1}
        &\mathcal{G}^{i,j} = CMA_1(\mathcal{V}_i, \mathcal{V}_j), \quad \mathcal{G}^{i,p} = CMA_1(\mathcal{V}_i, \mathcal{V}_p)\\
        &\textbf{G}_{i,j} = (\mathcal{V}_i, \mathcal{V}_j, \mathcal{G}^{i,j}), \quad \textbf{G}_{i,p} = (\mathcal{V}_i, \mathcal{V}_p, \mathcal{G}^{i,p}) \\
        &\overline{\mathcal{V}}_i = \parallel \{a(\textbf{G}_{i,j}), a(\textbf{G}_{i,p})\}, \quad a = MLP \circ CMA_2 \\
        &\mathcal{G}^{i,i} = MHSA_1(\overline{\mathcal{V}}_i) \\
        &\textbf{G}_{i,i} = (\overline{\mathcal{V}}_i, \mathcal{G}^{i,i}) \\
        &H_i = MLP \circ MHSA_2(\textbf{G}_{i,i})[-1] \\
        &\textbf{Repeat For Set} \quad \{j, p\} \quad \textbf{Then}\\
        &\mathcal{X}_m = f (\parallel \{H_i, H_j, H_p\})
    \end{aligned}
\end{equation}

Based on Definition \ref{definition1}, Lemma \ref{lemma1}, and Equation \ref{g_1}, we define MulTs as being composed of multiple subgraphs, with a series of complex function systems establishing hierarchical connections between these subgraphs. From the perspective of a single dominant-modality subgraph, it forms a tree. The integration of multiple dominant-modality trees ensembles a forest structure. In summary, we define MulTs as \textbf{Hierarchical Modal-wise Heterogeneous Graphs (HMHGs)}. Traditional forest structure of HMHG can be found in Figure \ref{fig:graph_structure_comp}.

\section{Motivation}

The aforementioned subgraphs can be transformed into a group of block-wise adjacency matrices and corresponding graphs as follows. 

\begin{equation}
    \begin{aligned}
        &\mathcal{G}_{inter}^{forward} = \begin{pmatrix}
            \mathcal{O}^{t,t} & \mathcal{G}^{t,v} & \mathcal{O}^{t,a} \\
            \mathcal{O}^{v,t} & \mathcal{O}^{v,v} & \mathcal{G}^{v,a} \\
            \mathcal{G}^{a,t} & \mathcal{O}^{a,v} & \mathcal{O}^{a,a} \\
          \end{pmatrix} \\ &\mathcal{G}_{inter}^{backward} = \begin{pmatrix}
            \mathcal{O}^{t,t} & \mathcal{O}^{t,v} & \mathcal{G}^{t,a} \\
            \mathcal{G}^{v,t} & \mathcal{O}^{v,v} & \mathcal{O}^{v,a} \\
            \mathcal{O}^{a,t} & \mathcal{G}^{a,v} & \mathcal{O}^{a,a} \\
          \end{pmatrix} \\
        \label{g_f}
        & \textbf{G}_{inter}^{forward} = (\mathcal{V}_m, \mathcal{G}_{inter}^{forward}) \\ &\textbf{G}_{inter}^{backward} = (\mathcal{V}_m, \mathcal{G}_{inter}^{backward})
    \end{aligned}    
\end{equation}

\begin{equation}
    \begin{aligned}
        &\mathcal{G}_{intra} = \begin{pmatrix}
        \mathcal{G}^{t,t} & \mathcal{O}^{t,v} & \mathcal{O}^{t,a} \\
        \mathcal{O}^{v,t} & \mathcal{G}^{v,v} & \mathcal{O}^{v,a} \\
        \mathcal{O}^{a,t} & \mathcal{O}^{a,v} & \mathcal{G}^{a,a} \\
        \end{pmatrix} \\
        \label{g_i}
        &\textbf{G}_{intra} = (\overline{\mathcal{V}}_m, \mathcal{G}_{intra})
    \end{aligned}    
\end{equation}

In the above equations, $\mathcal{O}^{i,j}$, where $\{i,j\}\in\{t,v,a\}$, refers to all zero matrix.

$\textbf{G}_{inter}^{forward}$ and $\textbf{G}_{inter}^{backward}$ in Equation \ref{g_f} are implemented for multimodal fusion, while $\textbf{G}_{intra}$ in Equation \ref{g_i} is for intra-modal enhancement.

This graph representation is mathematically equivalent to the traditional MulTs representation, which is an HMHG. However, it compresses the traditional forest structure into a single tree. Although it does not reduce the computational overhead regarding vertex information aggregation, it theoretically reduces the number of parameters to 1/3 of the traditional approach. 

Combined with $\textbf{G}_{inter}^{forward}$ and $\textbf{G}_{inter}^{backward}$, the overall multimodal fusion structure is composed of two opposing unidirectional cycle. They manage to make multimodal fusion complete without information disorder. Similarly, $\textbf{G}_{intra}$ also realizes complete intra-modal enhancement without information disorder. For more details about information disorder, please refer to Section \ref{inf_disorder}.

Realizing that this structure perfectly aligns with Insight \ref{insight1}, we are motivated to implement this idea and explore its potential benefits. 

\begin{figure}[t]
\centering
  \includegraphics[width=\linewidth]{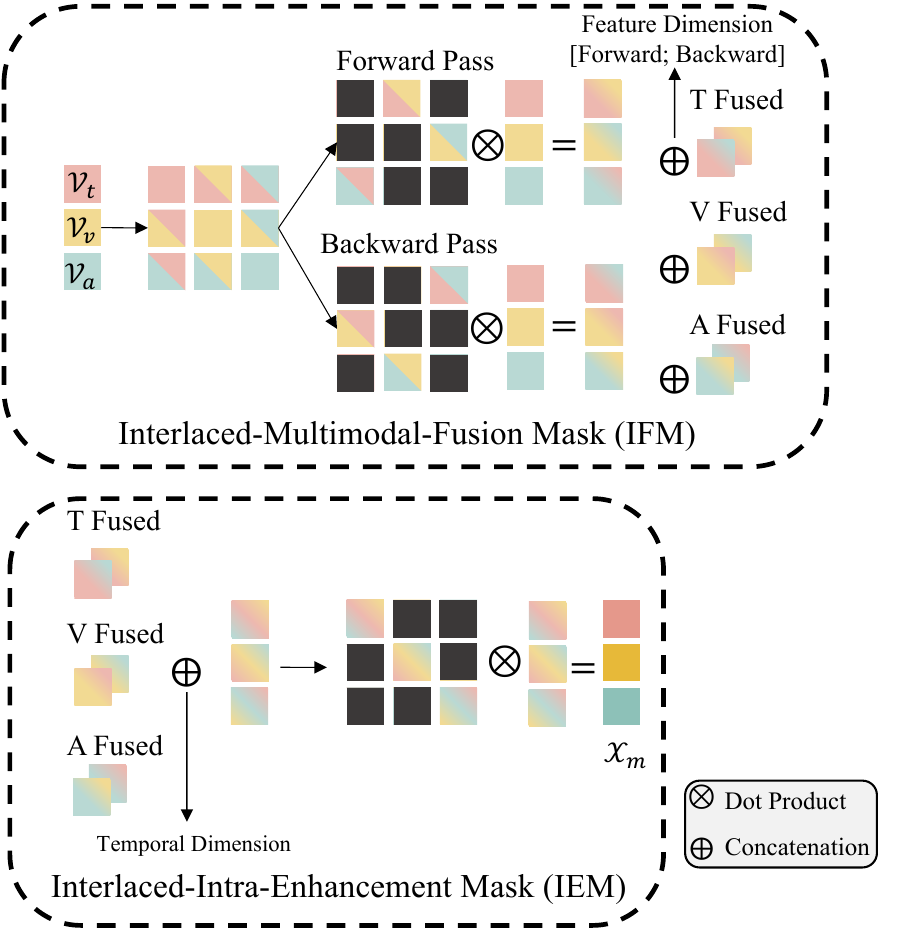}
  \caption{Interlaced Mask Mechanism. \textbf{Note}: detailed function system is omitted.}
    \label{fig:interlaced Mask}
\end{figure}

\section{All-Modal-In-One Fusion}

The core of the implementation of the graph structure defined in Equation \ref{g_f} and \ref{g_i} is a unique masking mechanism, which we call the \textbf{Interlaced Mask Mechanism (IM)}. There are two main parts in IM, \textbf{Interlaced-Multimodal-Fusion Mask (IFM)} and \textbf{Interlaced-Intra-Enhancement Mask (IEM)}.  Here, we define $\mathcal{J}^{i,j}$, where $\{i,j\}\in\{t,v,a\}$ refers to all negative infinity matrix.

\vspace{-8pt}
\begin{equation}
    \begin{aligned}
        \label{ifm}
        &\mathcal{M}_{inter}^{forward} = \begin{pmatrix}
            \mathcal{J}^{t,t} & \mathcal{O}^{t,v} & \mathcal{J}^{t,a} \\
            \mathcal{J}^{v,t} & \mathcal{J}^{v,v} & \mathcal{O}^{v,a} \\
            \mathcal{O}^{a,t} & \mathcal{J}^{a,v} & \mathcal{J}^{a,a} \\
          \end{pmatrix} \\ 
          &\mathcal{M}_{inter}^{backward} = \begin{pmatrix}
            \mathcal{J}^{t,t} & \mathcal{J}^{t,v} & \mathcal{O}^{t,a} \\
            \mathcal{O}^{v,t} & \mathcal{J}^{v,v} & \mathcal{J}^{v,a} \\
            \mathcal{J}^{a,t} & \mathcal{O}^{a,v} & \mathcal{J}^{a,a} \\
          \end{pmatrix}
    \end{aligned}
\end{equation}

\vspace{-8pt}
\begin{equation}
    \mathcal{M}_{intra} = \begin{pmatrix}
        \mathcal{O}^{t,t} & \mathcal{J}^{t,v} & \mathcal{J}^{t,a} \\
        \mathcal{J}^{v,t} & \mathcal{O}^{v,v} & \mathcal{J}^{v,a} \\
        \mathcal{J}^{a,t} & \mathcal{J}^{a,v} & \mathcal{O}^{a,a} \\
      \end{pmatrix}
\end{equation}

IFM contains $\mathcal{M}_{inter}^{forward}, \mathcal{M}_{inter}^{backward}$ in Equation \ref{ifm}, while $\mathcal{M}_{intra}$ is the IEM. 

Using IM, the graph defined in Equation \ref{g_f} and \ref{g_i} and its aggregation process can be easily defined.

\begin{equation}
    \begin{aligned}
        \label{allmodalinone fusion}
        &\mathcal{G}_{inter}^{forward} = MHSA_1(\mathcal{V}_m) + \mathcal{M}_{inter}^{forward} \\
        &\textbf{G}_{inter}^{forward} = (\mathcal{V}_m, \mathcal{G}_{inter}^{forward})\\ 
        &\mathcal{G}_{inter}^{backward} = MHSA_1(\mathcal{V}_m) + \mathcal{M}_{inter}^{backward} \\
        &\textbf{G}_{inter}^{backward} = (\mathcal{V}_m, \mathcal{G}_{inter}^{backward}) \\
        &a = MLP \circ MHSA_2 \\
        &\overline{\mathcal{V}}_m = \parallel\{a(\textbf{G}_{inter}^{forward}), a(\textbf{G}_{inter}^{backward})\}\\ 
        &\mathcal{G}_{intra} = MHSA_1(\overline{\mathcal{V}}_m) + \mathcal{M}_{intra} \\
        &\textbf{G}_{intra} = (\overline{\mathcal{V}}_m, \mathcal{G}_{intra}) \\ 
        &\mathcal{X}_m = f(\parallel Split(a(\textbf{G}_{intra}))[-1]) 
    \end{aligned}
\end{equation}

In the traditional approach, different subgraphs are decoupled and computed separately, with each having its own independent set of weights. However, based on the derived structure, the weights between these combinations can be shared. Specifically, the function system with 6 $CMA_{\{1,2\}}$, 3 $MHSA_{\{1,2\}}$, and 9 $MLP$ in MulTs is integrated to a function system of 3 $MHSA_{\{1,2\}}$ and 3 $MLP$. The computation visualization can be found in Figure \ref{fig:interlaced Mask}.

Due to the weight sharing strategy, we call this fusion method \textbf{All-Modal-In-One Fusion}. Based on this method and drawing inspiration from classical MulTs, we designed \textbf{Graph-Structured and Interlaced-Masked Multimodal Transformer (GsiT)}.

\section{Inner Decomposition for Efficiency}

The space complexity problem of GsiT as follows might be noticed. As $\mathcal{V}_m \in \mathbb{R}^{T_m \times d_m}$, where $T_m = T_t + T_v + T_a, d_m = d_{\{t,v,a\}}$, we assume that batch size $\mathcal{B}  \in \text{R}$. Although GsiT reduces the number of parameters to only 1/3 of MulT, in the runtime forward pass, the attention map of GsiT can achieve $O((T_m ^2) \times \mathcal{B}) = O((T_t + T_v + T_a) ^2 \times \mathcal{B})$. However, for MulTs, it is $O(T_i T_j \times \mathcal{B})$, where ${\{i, j\}}\in{\{t,v,a\}}$. Similarly, as for the adjacency matrix generation procedure. For GsiT, the computational complexity is $O((T_m ^2 d_m^f ) \times \mathcal{B})$, while for MulTs, it is $O(T_i T_j d_m^f \times \mathcal{B})$. The formal theoretical analysis of computational and space complexity is in Appendix \ref{complexity}.

Initially, it might seem that GsiT's complexity exceeds that of MulTs, which does not align with Insight \ref{insight1}. This issue can be easily resolved. After performing the shared qkv (query, key, value) projections on $\mathcal{V}_m$, we can decompose the sequences again according to their original lengths and apply internal operations according to the given IM. This approach ensures that the space complexity of the attention map remains the same, while that of static parameters is reduced to 1/3 of the theoretical value. This approach is called \textbf{Decomposition}, and we implement a simple block-sparse Triton kernel to optimize.

\section{Experiment}

\subsection{Experimental Setup}

We aim to check whether GsiT, its corresponding HMHG concept, and the IM Mechanism can be broadly applied to multiple models. To this end, we design comparative experiments between GsiT and several classic backbone models, and we embed not only GsiT itself but also its HMHG concept into multiple backbone-level models and MulTs in an appropriate manner to validate its broad effectiveness. We do not consider the parameters and computations of pre-trained language models in efficiency assessments, as these are consistent across all models. For further details on experimental settings, please refer to Appendix \ref{experiment_settings}.

\subsubsection{Datasets}

We evaluate GsiT and its HMHG concept on three widely used public datasets, CMU-MOSI  \citep{arxiv:mosi}, CMU-MOSEI  \citep{acl:mosei}, CH-SIMS \citep{acl:ch-sims}, and MIntRec \citep{mm:mintrec}. Please refer to Appendix \ref{datasets} for a more detailed description of datasets.

\subsubsection{Evaluation Criteria}

Following previous works \citep{acl:confede, ipm:cmhfm, taffc:mtmd}, several evaluation metrics are adopted. Binary classification accuracy (Acc-2), three classification accuracy (Acc-3), five classification accuracy (Acc-5), F1 Score (F1), seven classification accuracy (Acc-7), mean absolute error (MAE), and the correlation of the model's prediction with human (Corr).  Acc-2 and F1 are calculated in two ways: negative/non-negative(NN) and negative/positive(NP) on CMU-MOSI and CMU-MOSEI datasets. Acc-3 and Acc-5 are special metrics only for CH-SIMS. In MIntRec, Acc-20 refers to 20-class classification accuracy, Prec denotes precision, and Rec represents recall. Specifically, 'W' indicates the weighted result, introduced to address dataset imbalance. Additionally, for model efficiency, we provide the number of parameters Params (M), and floating-point operations per second FLOPS (G) to evaluate.

\begin{table*}[!htp]
  \caption{Comparison on CMU-MOSI and CMU-MOSEI. $\Delta$ denotes the numeric changes in metrics, $\dagger$ denotes that the results are reproduced, and w / denotes with. In particular, w / GsiT denotes simply adding GsiT into the original model, while w / HMHG denotes embedding the HMHG concept of GsiT into the original model.}
  \label{tab: Results_Compare_CMU-MOSI_CMU-MOSEI}
  \small
  \resizebox{1.0\linewidth}{!}
  {
    \begin{tabular}{l|ccccc|ccccc|cc}
      \toprule
      \multirow{2}{*}{Model} & \multicolumn{5}{c}{CMU-MOSI} & \multicolumn{5}{c}{CMU-MOSEI} &
      \multicolumn{2}{c}{Efficiency}\\
      & Acc-2(\%)$\uparrow$ & F1(\%)$\uparrow$ & Acc-7(\%)$\uparrow$ & MAE$\downarrow$ & Corr$\uparrow$ & Acc-2(\%)$\uparrow$ & F1(\%)$\uparrow$ & Acc-7(\%)$\uparrow$ & MAE$\downarrow$ & Corr$\uparrow$ & Params (M) $\downarrow$ & FLOPS (G) $\downarrow$ \\
      
      \midrule

      $\text{MulT}^\dagger$ & 79.6 / 81.4 & 79.1 / 81.0 & 36.2 & 0.923 & 0.686 & 78.1 / 83.7 & 78.9 / 83.7 & 53.4 & 0.559 & 0.740 & 5.251 &  26.294  \\

      $\textbf{GsiT}$ & \uuline{83.7} / \uuline{85.8} & \uuline{83.6} / \uuline{85.8} & \uuline{47.4} & \uuline{0.713} & \uuline{0.794} & \uuline{84.5}/ \uuline{85.6} & \uuline{84.4} / \uuline{85.2} & \uuline{54.1} & \uuline{0.536} & \uuline{0.764} & \uuline{1.695} & \uuline{26.224}  \\

      $\Delta$ & +4.1 / +4.4 & +4.5 / +4.8 & +11.2 & -0.210 & +0.108 & +6.4 / +1.9 & +5.5 / +1.5 & +0.7 & -0.023 & +0.024 & -67.7\% & -0.3\% \\ 

      \midrule

      $\text{Self-MM}^\dagger$ & 82.2 / 83.5 & 82.3 / 83.6 & 43.9 & 0.758 & 0.792 & 80.8 / 85.0 & 81.3 / 84.9 & 53.3 & 0.539 & 0.761 & \uuline{11.364} & \uuline{38.413} \\

      $\text{w/ GsiT}$ & \uuline{84.6} / \uuline{86.0} & \uuline{84.5} / \uuline{86.0} & \uuline{47.2} & \uuline{0.730} & 0.792 & \uuline{81.4} / \uuline{85.3} & \uuline{81.9} / \uuline{85.2} &  \uuline{54.1} & \uuline{0.536} & \uuline{0.762} & 13.059 & 64.637 \\
      $\Delta$ & +2.4 / +2.5 & +2.2 / +2.4 & +3.3 & -0.028 & - & +0.6 / +0.3 & +0.6 / +0.3 & +0.8 & -0.003 & +0.001  & +13.9\% & +68.3\% \\

      \midrule
      
      $\text{TETFN}^\dagger$ & 82.4 / 84.0	& 82.4 / 84.1	& 46.1	& 0.749 & 0.784 & 81.9 / 84.3 & 82.1 / 84.1 & \uuline{52.7} & \uuline{0.576} & 0.728 & 5.921 & 27.558 \\ 

      $\text{w/ HMHG}$ & \uuline{83.2} / \uuline{85.2} & \uuline{83.1} / \uuline{85.2}  &  \uuline{47.1} & \uuline{0.714} & \uuline{0.807} &  \uuline{84.6} / \uuline{84.8} & \uuline{84.5} / \uuline{84.5}  &  47.6 & 0.621 &  \uuline{0.749} & \uuline{2.365} & \uuline{27.488} \\
      
      $\Delta$ & +0.8 / +1.2 & +0.7 / +1.1 & +1.0 & -0.035 & +0.023 & +2.7 / +0.5 & +2.4 / +0.4 & -5.1 & -0.045 & +0.021 & -60.1\% & -0.3\% \\  

      \midrule

      $\text{ALMT}^\dagger$ &  82.1 / 83.3  & 82.1 / 83.3   & 45.5 &  0.730 & \uuline{0.791} & 81.4 / 83.5 & 81.6 / 83.3 & 49.2 & 0.583 & 0.731 &  2.604 & 19.876 \\

      $\text{w/ HMHG}$ & \uuline{83.2} / \uuline{84.6} & \uuline{83.1} / \uuline{84.5} &  \uuline{47.1} & \uuline{0.726} & 0.782 & \uuline{82.9} / \uuline{86.4} & \uuline{83.2} / \uuline{86.3} &  \uuline{51.5} & \uuline{0.541} & \uuline{0.773} & \uuline{2.506} & 19.876  \\
      $\Delta$ & +1.1 / +1.3 & +1.0 / +1.2 & +1.6 & -0.004 & -0.009 & +1.5 / +2.9 & +1.6 / +3.0 & +2.3 & -0.042 & +0.042 & -3.8\% & - \\
      \bottomrule
    \end{tabular}
  }
\end{table*}

\begin{table}[!htp]
  \caption{Additional comparison on CH-SIMS.}
  \label{tab: Results_Compare_SIMS}
  \resizebox{0.48\textwidth}{!}{
    \begin{tabular}{lcccccc}
      \toprule
      \multirow{2}{*}{Model} & \multicolumn{6}{c}{CH-SIMS} \\
      &Acc-2(\%)$\uparrow$ & Acc-3(\%)$\uparrow$ & Acc-5(\%)$\uparrow$ & F1(\%)$\uparrow$ & MAE$\downarrow$ & Corr$\uparrow$\\
      \midrule
      $\text{MulT}^\dagger$	&77.8	&65.3	&38.2	&77.7	&0.443	&0.578\\
      $\text{Self-MM}^\dagger$	&78.1 &65.2	&41.3	&78.2	&0.423	& 0.585\\
      $\text{TETFN}^\dagger$	&78.0	&64.4	&\uuline{42.9}	&78.0	& 0.425	& 0.582\\
      $\text{ALMT}^\dagger$	&77.2	&64.3	& 42.5	&77.6	& 0.419	& 0.581\\
      \textbf{GsiT} & \uuline{78.8} & \uuline{65.7} & 42.2 & \uuline{78.8} & \uuline{0.410} & \uuline{0.588} \\
      \bottomrule
    \end{tabular}
    }
\end{table}

\begin{table}[!htp]
  \caption{Additional comparison on MIntRec.}
  \label{tab: Results_Compare_MIntRec}
  \resizebox{0.48\textwidth}{!}{
    \begin{tabular}{lcccc}
      \toprule
      \multirow{2}{*}{Model} & \multicolumn{4}{c}{MIntRec} \\
      &Acc-20(\%)$\uparrow$ & F1 / F1-W(\%)$\uparrow$ & Prec / Prec-W(\%)$\uparrow$ & Rec / Rec-W(\%)$\uparrow$ \\
      \midrule
      $\text{MulT}^\dagger$	&71.2	&68.2 / 71.1	&68.9 / 71.4	&68.1 / 71.2\\
      $\text{MMIM}^\dagger$	&70.8	&68.7 / 71.0	&69.2 / 71.8	&68.9 / 70.8\\
      $\textbf{GsiT}$	&\uuline{72.6}	&\uuline{69.4} / \uuline{72.7}	&\uuline{69.4} / \uuline{73.5}	&\uuline{70.1} / \uuline{72.6}\\
      \bottomrule
    \end{tabular}
    }
\end{table}

\subsubsection{Baseline Models}

\footnotetext[1]{\url{https://github.com/thuiar/MMSA} \label{mmsa}}

To clarify our approach, in our concept, the core multimodal fusion module and the learning framework in MSA are recognized as backbone-level models. MulT \citep{acl:mult}, Self-MM \citep{aaai:self-mm}, TETFN \citep{pr:tetfn}, and ALMT \citep{emnlp:almt} are selected as the baseline models for comparison. For further evaluation on MIntRec, we also incorporate MMIM \citep{emnlp:mmim}.

MulT and Self-MM are widely adopted backbone-level models, whereas TETFN combines elements of both MulT and Self-MM within a text-oriented framework, serving as a pure MulTs-based model. ALMT, on the other hand, builds upon the concepts of MulT and attention bottleneck, evolving into a next-generation MulTs-like architecture. The source code for these baselines is available on the GitHub page\textsuperscript{\ref{mmsa}}, with detailed introductions provided in Appendix \ref{baselines}.

In our experimental setup, we use MulT as the primary baseline for performance comparison due to its foundational role in MulTs-based models. Additionally, we integrate GsiT with Self-MM, one of the most prevalent self-supervised learning frameworks in MSA, to evaluate its effectiveness. Furthermore, we embed HMHG into both TETFN and ALMT—representative MulTs-based and MulTs-like models—to validate its enhancement capabilities.

\subsection{Results}

In all tables, double-underline denotes the superior performance, $\uparrow$ denotes that higher is better while $\downarrow$ denotes the opposite.

\subsubsection{Main Results}

The main results of the experiment are shown in \ref{tab: Results_Compare_CMU-MOSI_CMU-MOSEI}.

Compared with MulT, GsiT significantly outperforms MulT across all metrics while having substantially fewer parameters without additional computational overhead. This observation also holds for the other baseline models in our comparison. 

This demonstrates that GsiT and its HMHG concept are effective in enhancing performance across a variety of models. Firstly, as a standalone model, GsiT already exhibits impressive performance. Secondly, when integrated as a module into the classic self-supervised learning framework Self-MM, it notably improves overall performance. Additionally, replacing the core fusion framework of the MulTs-based model TETFN with the HMHG form results in significant improvements in both performance and efficiency. Finally, modifying the core AHL module of the MulTs-based architecture ALMT to the HMHG form also leads to a marked enhancement in performance.

Regarding the efficiency drop observed when integrating GsiT into Self-MM, it is important to note that Self-MM, as a self-supervised learning framework, primarily employs simple linear layers for multimodal fusion. Consequently, the addition of GsiT introduces more complex components, leading to an expected and reasonable decrease in efficiency.

The Acc-7 in TETFN significantly dropped after embedding HMHG. This is attributed to the modification of the IFM to accommodate the TET module, as defined in Equation \ref{structure3}, rather than following our initial design in Equation \ref{ifm}. Although this change maintained information integrity, it resulted in repeated bi-modality combinations within a single Encoder, limiting the model's ability to effectively integrate multimodal information. For more details, see Section \ref{ablationstudy}.

The modest efficiency improvement in ALMT can be attributed to that it is not a pure MulT-based model (TETFN is a pure MulT-based model). The relatively small scale of its AHL module has a small impact on the overall model's computational overhead. Nevertheless, the performance gains achieved by incorporating HMHG still demonstrate the significant benefits of the weight-sharing scheme provided by the IM mechanism.

The additional experiment on the Chinese dataset CH-SIMS, as shown in Table \ref{tab: Results_Compare_SIMS}, highlights GsiT's superior performance. In this backbone-level model comparison, GsiT outperforms both naive Self-MM and naive MulT across all metrics, and surpasses ALMT in most of the metrics. Furthermore, when compared with TETFN, which integrates Self-MM and MulT, GsiT demonstrates its advanced capabilities in most of the metrics. This underscores GsiT's next-level performance as a backbone multimodal fusion model. Additionally, these results confirm GsiT's robust multilingual capabilities.

Also, the extended experiment on the multimodal intent recognition dataset MIntRec, as shown in Table \ref{tab: Results_Compare_MIntRec}, highlights GsiT's superior performance. GsiT outperforms MulT and MMIM across all metrics, demonstrating its strong generalization capability in broader multimodal domains. 

\begin{table}[!t]
  \caption{Ablation Study on CMU-MOSI for GsiT.}
  \label{tab: Ablation Study}
  \resizebox{\linewidth}{!}  
  {
    \begin{tabular}{lcccccc}
      \toprule
      \multirow{2}{*}{Description} & \multicolumn{5}{c}{CMU-MOSI} \\
       & Acc-2$\uparrow$ & F1$\uparrow$ & Acc-7$\uparrow$ & MAE$\downarrow$ & Corr$\uparrow$ \\
      \midrule
      \text{Orginal} & \uuline{83.7} / \uuline{85.8} & \uuline{83.6} / \uuline{85.8} & \uuline{47.4} & \uuline{0.713} & 0.794 \\
      \text{Structure-1} & 83.5 / 85.5 & 83.4 / 85.4 & 46.5 & 0.721 & \uuline{0.798} \\
      \text{Structure-2} & 83.2 / 84.9 & 83.2 / 84.9 & 43.8 & 0.729 & 0.796 \\
      \text{Structure-3} & 83.4 / 85.2 & 83.3 / 85.2 & 45.5 & 0.726 & 0.783 \\
      \text{Self-Only} & 82.5 / 84.2 & 82.5 / 84.2 & 45.5 & 0.734 & 0.793 \\
      \bottomrule
    \end{tabular}
  }
\end{table}

\subsubsection{Ablation Study}

\label{ablationstudy}

In this section, we primarily explore the structure of the Interlaced Fusion Mask (IFM) to investigate how different graph structures impact the performance of the GsiT architecture. At this point, the \textit{Original Structure} and \textit{Structures 1 to 3} all adhere to the basic paradigm of HMHG, ensuring that multimodal information remains coherent. In contrast, the \textit{Self-Only} variant leads to information disorder. Here, we define the information fusion operation from $\mathcal{V}_j$ to $\mathcal{V}_i$ is as $j \rightarrow i$, where $\{i,j\} \in \{t,v,a\}$. Detailed illustration to information disorder can be found in Section \ref{inf_disorder}. 

In our concept, each set of bi-modality combination subgraphs has non-overlapping dominant modalities, and the combination of dominant and auxiliary modalities does not repeat regardless of the order. For instance, the same set will not simultaneously contain $\textbf{G}_{t,v}$ and $\textbf{G}_{a,v}$ to prevent the disorder of temporal sequence information across modality sequences. Similarly, the same set will not simultaneously contain $\textbf{G}_{t,v}$ and $\textbf{G}_{v,t}$, even if their dominant modalities are different, to ensure that each set's corresponding module learns the fusion information of the most diverse combinations, thereby enhancing the fusion performance. Thus, we designed the \textit{Original Structure}.

As shown in Table \ref{tab: Ablation Study}, the original structure is superior to the other three theoretically feasible structures in most of the metrics, which aligns with our concept. The four theoretically feasible structures are superior to the self-only structure, which is theoretically infeasible and causes information disorder.

\textit{Original Structure}: The original structure is defined as two opposing unidirectional ring graphs. They both realize cyclic all-modal-in-one fusion, which makes trimodal information fully interact in shared model weights. The structure is: $\{t \rightarrow v, v \rightarrow a, a \rightarrow t\}$, $\{a \rightarrow v, v \rightarrow t, t \rightarrow a \}$. 

\textit{Structure-1}: Structure-1 realizes all-modal-in-one fusion, but the information passing is not cyclic. The structure is: $\{a \rightarrow v, v \rightarrow a, a \rightarrow t\}$, $\{v \rightarrow t, t \rightarrow v, t \rightarrow a\}$. 

\textit{Structure-2}: Structure-2 realizes all-modal-in-one fusion, but the information passing is not cyclic. The structure is: $\{v \rightarrow t, t \rightarrow v, v \rightarrow a\}$, $\{a \rightarrow t, t \rightarrow a, a \rightarrow v\}$. 

\textit{Structure-3}: Structure-3 realizes all-modal-in-one fusion, but the information passing is not cyclic. The structure is: $\{a \rightarrow v, v \rightarrow a, v\rightarrow t\}$, $\{a \rightarrow t, t \rightarrow a, t \rightarrow v\}$. 


\textit{Self-Only}: Self-Only mask only contains masks intra-modal subgraphs. The structure is $\{t \rightarrow t, v \rightarrow v, a \rightarrow a\}$.

For more specific representations of the aforementioned structures, please refer to Appendix \ref{graph_structure}.


\section{Further Analysis}

\subsection{Information Disorder}
\label{inf_disorder}

Take $\mathcal{G}_{inter}^{forward}$ as an example.

\vspace{-8pt}
\begin{equation}
\begin{aligned}
    &\mathcal{G}_{inter}^{forward} = \begin{pmatrix}
        \mathcal{O}^{t,t} & \mathcal{G}_d^{t,v} & \mathcal{O}^{t,a} \\
        \mathcal{O}^{v,t} & \mathcal{O}^{v,v} & \mathcal{G}_d^{v,a} \\
        \mathcal{G}_d^{a,t} & \mathcal{O}^{a,v} & \mathcal{O}^{a,a} \\
      \end{pmatrix} \\ &\mathcal{G}_{inter}^{forward'} = \begin{pmatrix}
        \mathcal{O}^{t,t} & \mathcal{G}_d^{t,v'} & \mathcal{G}_d^{t,a'} \\
        \mathcal{O}^{v,t} & \mathcal{O}^{v,v} & \mathcal{G}_d^{v,a} \\
        \mathcal{G}_d^{a,t} & \mathcal{O}^{a,v} & \mathcal{O}^{a,a} \\
      \end{pmatrix}
\end{aligned}
\end{equation}

It is important to note that $\mathcal{G}_d^{t,v} \neq \mathcal{G}_d^{t,v'}$. The reason lies in Equation \ref{softmax_staff}, taking the first row block as an example.

\vspace{-8pt}
\begin{equation}
    \label{softmax_staff}
    \begin{aligned}
        &\begin{pmatrix}
            \mathcal{O}^{t,t} & \mathcal{G}_d^{t,v} & \mathcal{O}^{t,a} 
        \end{pmatrix} = \mathcal{S} \circ \mathcal{D}
        \begin{pmatrix}
            \mathcal{J}^{t,t} & \mathcal{E}^{t,v} & \mathcal{J}^{t,a} 
        \end{pmatrix} \\ 
        &\begin{pmatrix}
            \mathcal{O}^{t,t} & \mathcal{G}_d^{t,v'} & \mathcal{G}_d^{t,a'}
        \end{pmatrix} = \mathcal{S} \circ \mathcal{D}
        \begin{pmatrix}
            \mathcal{J}^{t,t} & \mathcal{E}^{t,v} & \mathcal{E}^{t,a} 
        \end{pmatrix} 
    \end{aligned}
\end{equation} 

The softmax function operates on a row-wise basis, converting values to probabilities. Therefore, if row elements include subgraphs other than the required modal subgraphs, it will affect the probability distribution of the desired subgraphs, thus affecting the result and causing information disorder.

For the original $\mathcal{G}_{inter}^{forward}$, the fusion process is as follows:

\vspace{-8pt}
\begin{equation}
\begin{aligned}
    \overline{\mathcal{V}}_m^{forward} 
    &= \mathcal{G}_{inter}^{forward} \mathcal{W}_v \mathcal{V}_m \\
    & = \begin{pmatrix} 
        \mathcal{O}^{t,t} & \mathcal{G}_d^{t,v} & \mathcal{O}^{t,a} \\
        \mathcal{O}^{v,t} & \mathcal{O}^{v,v} & \mathcal{G}_d^{v,a} \\
        \mathcal{G}_d^{a,t} & \mathcal{O}^{a,v} & \mathcal{O}^{a,a} \\
      \end{pmatrix} \cdot
      \begin{pmatrix}
      \mathcal{W}_v \mathcal{V}_t \\
      \mathcal{W}_v \mathcal{V}_v \\
      \mathcal{W}_v \mathcal{V}_a \\
      \end{pmatrix} \\
      &= \begin{pmatrix}
      \mathcal{G}_d^{a,t} \mathcal{W}_v \mathcal{V}_t \\
      \mathcal{G}_d^{t,v} \mathcal{W}_v \mathcal{V}_v \\
      \mathcal{G}_d^{v,a}\mathcal{W}_v \mathcal{V}_a \\
      \end{pmatrix}
\end{aligned}
\end{equation}

However, for the modified version $\mathcal{G}_{inter}^{forward'}$.

\vspace{-8pt}
\begin{equation}
    \begin{aligned}
        \overline{\mathcal{V}}_m^{forward'} &= \mathcal{G}_{inter}^{forward'} \mathcal{W}_v \mathcal{V}_m \\ 
        &= \begin{pmatrix}
            \mathcal{O}^{t,t} & \mathcal{G}_d^{t,v'} & \mathcal{G}_d^{t,a'} \\
            \mathcal{O}^{v,t} & \mathcal{O}^{v,v} & \mathcal{G}_d^{v,a} \\
            \mathcal{G}_d^{a,t} & \mathcal{O}^{a,v} & \mathcal{O}^{a,a} \\
          \end{pmatrix} \cdot
          \begin{pmatrix}
          \mathcal{W}_v \mathcal{V}_t \\
          \mathcal{W}_v \mathcal{V}_v \\
          \mathcal{W}_v \mathcal{V}_a \\
          \end{pmatrix} \\ 
          &= \begin{pmatrix}
          \mathcal{G}_d^{a,t} \mathcal{W}_v \mathcal{V}_t \\
          \mathcal{G}_d^{t,v} \mathcal{W}_v \mathcal{V}_v \\
          (\mathcal{G}_d^{v,a'} + \mathcal{G}_d^{t,a'})\mathcal{W}_v \mathcal{V}_a \\
          \end{pmatrix}
    \end{aligned}
\end{equation}

This is the cause of information disorder.

\begin{figure}[t]
\centering
  \includegraphics[width=\linewidth]{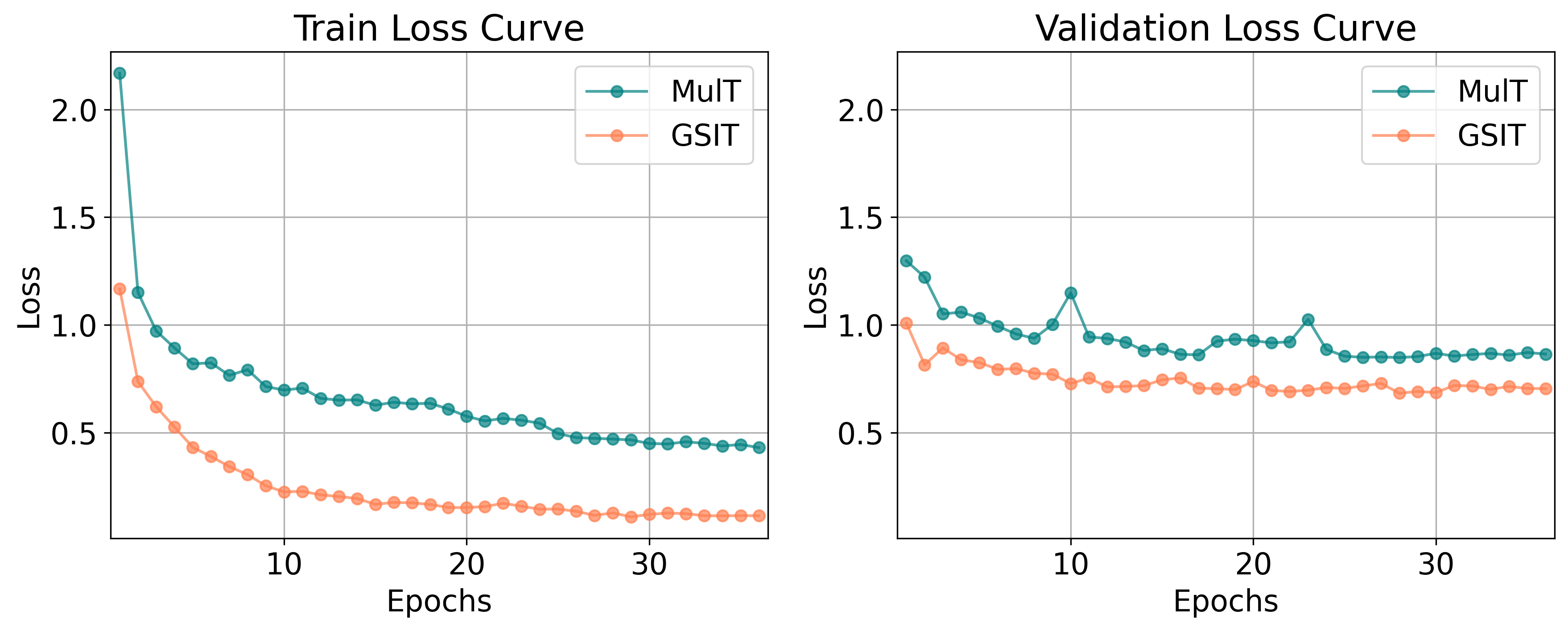}
  \caption{The loss curve of MulT and GsiT training phase on CMU-MOSI.}
    \label{fig:loss curve}
\end{figure}

\subsection{Convergence Analysis}

To evaluate convergence, we collected loss decline curves for MulT and GsiT using early stopping (8 times). As shown in Figure \ref{fig:loss curve}, both models converge in the same training rounds, but GsiT achieves consistently lower loss values in all training and validation phases, demonstrating superior convergence and loss optimization capabilities.

\begin{table}[!t]
  \caption{Weight Distribution.}
  \label{tab: Weight Distribution}
  \resizebox{\linewidth}{!}  
  {
    \begin{tabular}{lcccc}
      \toprule
      Model & Mean & Variance & Skewness & Kurtosis \\
      \midrule
      \text{GsiT} & -0.0001 & 0.0027 & 0.0016 & -0.7616 \\
      \text{MulT} & 0.0000 & 0.0032 & 0.0004 & -0.8505  \\
      \bottomrule
    \end{tabular}
  }
\end{table}

\subsection{Weight Distribution}

As shown in Table \ref{tab: Weight Distribution}, GsiT and MulT have means near zero, minimizing initial biases. GsiT's lower variance indicates concentrated weights, reduced noise sensitivity, and lower overfitting risk. Both models show near-zero skewness, reflecting symmetry and stability. GsiT's kurtosis, closer to zero than MulT's, suggests better extreme value control and a superior adaptability-stability balance, enhancing training and generalization.

\section{Conclusion}

This work uncovers that MulTs are essentially hierarchical modal-wise heterogeneous graphs (HMHGs). Leveraging this theorem, we propose an Interlaced Mask (IM) mechanism to develop the Graph-Structured Interlaced-Masked Multimodal Transformer (GsiT). GsiT, formally equivalent to MulTs, achieves efficient weight sharing without information disorder, enabling All-Modal-In-One fusion using just 1/3 the parameters of conventional MulTs and outperforms them significantly. Decomposition is implemented to make sure this without additional computational overhead. Experiments on popular MSA datasets, including integrating GsiT and HMHG into several state-of-the-art models, demonstrate notable performance and efficiency improvements. 

\section*{Limitations}

While our proposed GsiT and HMHG concept has shown promising results in multimodal sentiment analysis, there are some limitations to consider. Firstly, our method is designed for multimodal sentiment analysis only, without considering other multimodal tasks. The performance of the model when one of these modalities is missing is not considered. Additionally, in the first level of HMHG and GsiT, which is the multimodal fusion encoder-pair, we have not utilized representation learning methods such as contrastive learning to enhance the representation of the fused information in both directions, which is a direction worth exploring in the future.

\normalem 
\bibliography{custom}

\begin{thebibliography}{37}
\providecommand{\natexlab}[1]{#1}

\bibitem[{Baevski et~al.(2020)Baevski, Zhou, Mohamed, and Auli}]{neurips:wav2vec2}
Alexei Baevski, Yuhao Zhou, Abdelrahman Mohamed, and Michael Auli. 2020.
\newblock wav2vec 2.0: {A} framework for self-supervised learning of speech representations.
\newblock In \emph{Advances in Neural Information Processing Systems 33: Annual Conference on Neural Information Processing Systems 2020, NeurIPS 2020, December 6-12, 2020, virtual}.

\bibitem[{Bagher~Zadeh et~al.(2018)Bagher~Zadeh, Liang, Poria, Cambria, and Morency}]{acl:mosei}
AmirAli Bagher~Zadeh, Paul~Pu Liang, Soujanya Poria, Erik Cambria, and Louis-Philippe Morency. 2018.
\newblock \href {https://doi.org/10.18653/v1/P18-1208} {Multimodal language analysis in the wild: {CMU}-{MOSEI} dataset and interpretable dynamic fusion graph}.
\newblock In \emph{Proceedings of the 56th Annual Meeting of the Association for Computational Linguistics (Volume 1: Long Papers)}, pages 2236--2246, Melbourne, Australia. Association for Computational Linguistics.

\bibitem[{Baltrusaitis et~al.(2016)Baltrusaitis, Robinson, and Morency}]{wacv:openface}
Tadas Baltrusaitis, Peter Robinson, and Louis{-}Philippe Morency. 2016.
\newblock Openface: An open source facial behavior analysis toolkit.
\newblock In \emph{2016 {IEEE} Winter Conference on Applications of Computer Vision, {WACV} 2016, Lake Placid, NY, USA, March 7-10, 2016}, pages 1--10. {IEEE} Computer Society.

\bibitem[{Baltrusaitis et~al.(2018)Baltrusaitis, Zadeh, Lim, and Morency}]{fg:openface2}
Tadas Baltrusaitis, Amir Zadeh, Yao~Chong Lim, and Louis{-}Philippe Morency. 2018.
\newblock Openface 2.0: Facial behavior analysis toolkit.
\newblock In \emph{13th {IEEE} International Conference on Automatic Face {\&} Gesture Recognition, {FG} 2018, Xi'an, China, May 15-19, 2018}, pages 59--66. {IEEE} Computer Society.

\bibitem[{Brody et~al.(2022)Brody, Alon, and Yahav}]{iclr:gatv2}
Shaked Brody, Uri Alon, and Eran Yahav. 2022.
\newblock How attentive are graph attention networks?
\newblock In \emph{The Tenth International Conference on Learning Representations, {ICLR} 2022, Virtual Event, April 25-29, 2022}. OpenReview.net.

\bibitem[{Degottex et~al.(2014)Degottex, Kane, Drugman, Raitio, and Scherer}]{icassp:covarep}
Gilles Degottex, John Kane, Thomas Drugman, Tuomo Raitio, and Stefan Scherer. 2014.
\newblock {COVAREP} - {A} collaborative voice analysis repository for speech technologies.
\newblock In \emph{{IEEE} International Conference on Acoustics, Speech and Signal Processing, {ICASSP} 2014, Florence, Italy, May 4-9, 2014}, pages 960--964. {IEEE}.

\bibitem[{Devlin et~al.(2019)Devlin, Chang, Lee, and Toutanova}]{naacl:bert}
Jacob Devlin, Ming-Wei Chang, Kenton Lee, and Kristina Toutanova. 2019.
\newblock \href {https://doi.org/10.18653/v1/N19-1423} {{BERT}: Pre-training of deep bidirectional transformers for language understanding}.
\newblock In \emph{Proceedings of the 2019 Conference of the North {A}merican Chapter of the Association for Computational Linguistics: Human Language Technologies, Volume 1 (Long and Short Papers)}, pages 4171--4186, Minneapolis, Minnesota. Association for Computational Linguistics.

\bibitem[{Gandhi et~al.(2023)Gandhi, Adhvaryu, Poria, Cambria, and Hussain}]{if:survey}
Ankita Gandhi, Kinjal Adhvaryu, Soujanya Poria, Erik Cambria, and Amir Hussain. 2023.
\newblock Multimodal sentiment analysis: {A} systematic review of history, datasets, multimodal fusion methods, applications, challenges and future directions.
\newblock \emph{Inf. Fusion}, 91:424--444.

\bibitem[{Han et~al.(2021)Han, Chen, and Poria}]{emnlp:mmim}
Wei Han, Hui Chen, and Soujanya Poria. 2021.
\newblock \href {https://doi.org/10.18653/v1/2021.emnlp-main.723} {Improving multimodal fusion with hierarchical mutual information maximization for multimodal sentiment analysis}.
\newblock In \emph{Proceedings of the 2021 Conference on Empirical Methods in Natural Language Processing}, pages 9180--9192, Online and Punta Cana, Dominican Republic. Association for Computational Linguistics.

\bibitem[{He et~al.(2016)He, Zhang, Ren, and Sun}]{cvpr:resnet}
Kaiming He, Xiangyu Zhang, Shaoqing Ren, and Jian Sun. 2016.
\newblock \href {https://doi.org/10.1109/CVPR.2016.90} {Deep residual learning for image recognition}.
\newblock In \emph{2016 {IEEE} Conference on Computer Vision and Pattern Recognition, {CVPR} 2016, Las Vegas, NV, USA, June 27-30, 2016}, pages 770--778. {IEEE} Computer Society.

\bibitem[{Hochreiter(1997)}]{nc:lstm}
S~Hochreiter. 1997.
\newblock Long short-term memory.
\newblock \emph{Neural Computation MIT-Press}.

\bibitem[{Lin and Hu(2024)}]{taffc:mtmd}
Ronghao Lin and Haifeng Hu. 2024.
\newblock Multi-task momentum distillation for multimodal sentiment analysis.
\newblock \emph{{IEEE} Trans. Affect. Comput.}, 15(2):549--565.

\bibitem[{Liu et~al.(2018)Liu, Shen, Lakshminarasimhan, Liang, Bagher~Zadeh, and Morency}]{acl:lmf}
Zhun Liu, Ying Shen, Varun~Bharadhwaj Lakshminarasimhan, Paul~Pu Liang, AmirAli Bagher~Zadeh, and Louis-Philippe Morency. 2018.
\newblock \href {https://doi.org/10.18653/v1/P18-1209} {Efficient low-rank multimodal fusion with modality-specific factors}.
\newblock In \emph{Proceedings of the 56th Annual Meeting of the Association for Computational Linguistics (Volume 1: Long Papers)}, pages 2247--2256, Melbourne, Australia. Association for Computational Linguistics.

\bibitem[{Mai et~al.(2023{\natexlab{a}})Mai, Xing, He, Zeng, and Hu}]{acmtmm:gpfn}
Sijie Mai, Songlong Xing, Jiaxuan He, Ying Zeng, and Haifeng Hu. 2023{\natexlab{a}}.
\newblock \href {https://doi.org/10.1145/3542927} {Multimodal graph for unaligned multimodal sequence analysis via graph convolution and graph pooling}.
\newblock \emph{{ACM} Trans. Multim. Comput. Commun. Appl.}, 19(2):54:1--54:24.

\bibitem[{Mai et~al.(2023{\natexlab{b}})Mai, Zeng, Zheng, and Hu}]{taffc:hycon}
Sijie Mai, Ying Zeng, Shuangjia Zheng, and Haifeng Hu. 2023{\natexlab{b}}.
\newblock Hybrid contrastive learning of tri-modal representation for multimodal sentiment analysis.
\newblock \emph{{IEEE} Trans. Affect. Comput.}, 14(3):2276--2289.

\bibitem[{Peng et~al.(2023)Peng, Wu, Zhang, Cheng, Tan, Yi, and Huang}]{eswa:fmlmsn}
Junjie Peng, Ting Wu, Wenqiang Zhang, Feng Cheng, Shuhua Tan, Fen Yi, and Yansong Huang. 2023.
\newblock A fine-grained modal label-based multi-stage network for multimodal sentiment analysis.
\newblock \emph{Expert Syst. Appl.}, 221:119721.

\bibitem[{Rahman et~al.(2020)Rahman, Hasan, Lee, Bagher~Zadeh, Mao, Morency, and Hoque}]{acl:mag-bert}
Wasifur Rahman, Md~Kamrul Hasan, Sangwu Lee, AmirAli Bagher~Zadeh, Chengfeng Mao, Louis-Philippe Morency, and Ehsan Hoque. 2020.
\newblock \href {https://doi.org/10.18653/v1/2020.acl-main.214} {Integrating multimodal information in large pretrained transformers}.
\newblock In \emph{Proceedings of the 58th Annual Meeting of the Association for Computational Linguistics}, pages 2359--2369, Online. Association for Computational Linguistics.

\bibitem[{Tsai et~al.(2019{\natexlab{a}})Tsai, Bai, Liang, Kolter, Morency, and Salakhutdinov}]{acl:mult}
Yao-Hung~Hubert Tsai, Shaojie Bai, Paul~Pu Liang, J.~Zico Kolter, Louis-Philippe Morency, and Ruslan Salakhutdinov. 2019{\natexlab{a}}.
\newblock \href {https://doi.org/10.18653/v1/P19-1656} {Multimodal transformer for unaligned multimodal language sequences}.
\newblock In \emph{Proceedings of the 57th Annual Meeting of the Association for Computational Linguistics}, pages 6558--6569, Florence, Italy. Association for Computational Linguistics.

\bibitem[{Tsai et~al.(2019{\natexlab{b}})Tsai, Liang, Zadeh, Morency, and Salakhutdinov}]{iclr:mfm}
Yao{-}Hung~Hubert Tsai, Paul~Pu Liang, Amir Zadeh, Louis{-}Philippe Morency, and Ruslan Salakhutdinov. 2019{\natexlab{b}}.
\newblock Learning factorized multimodal representations.
\newblock In \emph{7th International Conference on Learning Representations, {ICLR} 2019, New Orleans, LA, USA, May 6-9, 2019}. OpenReview.net.

\bibitem[{Vaswani et~al.(2017)Vaswani, Shazeer, Parmar, Uszkoreit, Jones, Gomez, Kaiser, and Polosukhin}]{nips:transformer}
Ashish Vaswani, Noam Shazeer, Niki Parmar, Jakob Uszkoreit, Llion Jones, Aidan~N. Gomez, Lukasz Kaiser, and Illia Polosukhin. 2017.
\newblock Attention is all you need.
\newblock In \emph{Advances in Neural Information Processing Systems 30: Annual Conference on Neural Information Processing Systems 2017, December 4-9, 2017, Long Beach, CA, {USA}}, pages 5998--6008.

\bibitem[{Velickovic et~al.(2018)Velickovic, Cucurull, Casanova, Romero, Li{\`{o}}, and Bengio}]{iclr:gat}
Petar Velickovic, Guillem Cucurull, Arantxa Casanova, Adriana Romero, Pietro Li{\`{o}}, and Yoshua Bengio. 2018.
\newblock Graph attention networks.
\newblock In \emph{6th International Conference on Learning Representations, {ICLR} 2018, Vancouver, BC, Canada, April 30 - May 3, 2018, Conference Track Proceedings}. OpenReview.net.

\bibitem[{Wang et~al.(2023{\natexlab{a}})Wang, Guo, Tian, Liu, He, and Luo}]{pr:tetfn}
Di~Wang, Xutong Guo, Yumin Tian, Jinhui Liu, Lihuo He, and Xuemei Luo. 2023{\natexlab{a}}.
\newblock {TETFN:} {A} text enhanced transformer fusion network for multimodal sentiment analysis.
\newblock \emph{Pattern Recognit.}, 136:109259.

\bibitem[{Wang et~al.(2023{\natexlab{b}})Wang, Liu, Wang, Tian, He, and Gao}]{tmm:cenet}
Di~Wang, Shuai Liu, Quan Wang, Yumin Tian, Lihuo He, and Xinbo Gao. 2023{\natexlab{b}}.
\newblock Cross-modal enhancement network for multimodal sentiment analysis.
\newblock \emph{{IEEE} Trans. Multim.}, 25:4909--4921.

\bibitem[{Wang et~al.(2024)Wang, Peng, Zheng, Zhao, and Zhu}]{ipm:cmhfm}
Lan Wang, Junjie Peng, Cangzhi Zheng, Tong Zhao, and Li'an Zhu. 2024.
\newblock A cross modal hierarchical fusion multimodal sentiment analysis method based on multi-task learning.
\newblock \emph{Inf. Process. Manag.}, 61(2):103675.

\bibitem[{Wu et~al.(2024)Wu, Gong, Koo, and Hirschberg}]{naacl:mmml}
Zehui Wu, Ziwei Gong, Jaywon Koo, and Julia Hirschberg. 2024.
\newblock \href {https://doi.org/10.18653/v1/2024.naacl-long.197} {Multimodal multi-loss fusion network for sentiment analysis}.
\newblock In \emph{Proceedings of the 2024 Conference of the North American Chapter of the Association for Computational Linguistics: Human Language Technologies (Volume 1: Long Papers)}, pages 3588--3602, Mexico City, Mexico. Association for Computational Linguistics.

\bibitem[{Yang et~al.(2021)Yang, Wang, Yi, Zhu, Rehman, Zadeh, Poria, and Morency}]{naacl:mtag}
Jianing Yang, Yongxin Wang, Ruitao Yi, Yuying Zhu, Azaan Rehman, Amir Zadeh, Soujanya Poria, and Louis-Philippe Morency. 2021.
\newblock \href {https://doi.org/10.18653/v1/2021.naacl-main.79} {{MTAG}: Modal-temporal attention graph for unaligned human multimodal language sequences}.
\newblock In \emph{Proceedings of the 2021 Conference of the North American Chapter of the Association for Computational Linguistics: Human Language Technologies}, pages 1009--1021, Online. Association for Computational Linguistics.

\bibitem[{Yang et~al.(2023)Yang, Yu, Niu, Guo, and Xu}]{acl:confede}
Jiuding Yang, Yakun Yu, Di~Niu, Weidong Guo, and Yu~Xu. 2023.
\newblock \href {https://doi.org/10.18653/v1/2023.acl-long.421} {{C}on{FEDE}: Contrastive feature decomposition for multimodal sentiment analysis}.
\newblock In \emph{Proceedings of the 61st Annual Meeting of the Association for Computational Linguistics (Volume 1: Long Papers)}, pages 7617--7630, Toronto, Canada. Association for Computational Linguistics.

\bibitem[{Yu et~al.(2020)Yu, Xu, Meng, Zhu, Ma, Wu, Zou, and Yang}]{acl:ch-sims}
Wenmeng Yu, Hua Xu, Fanyang Meng, Yilin Zhu, Yixiao Ma, Jiele Wu, Jiyun Zou, and Kaicheng Yang. 2020.
\newblock \href {https://doi.org/10.18653/v1/2020.acl-main.343} {{CH}-{SIMS}: A {C}hinese multimodal sentiment analysis dataset with fine-grained annotation of modality}.
\newblock In \emph{Proceedings of the 58th Annual Meeting of the Association for Computational Linguistics}, pages 3718--3727, Online. Association for Computational Linguistics.

\bibitem[{Yu et~al.(2021)Yu, Xu, Yuan, and Wu}]{aaai:self-mm}
Wenmeng Yu, Hua Xu, Ziqi Yuan, and Jiele Wu. 2021.
\newblock Learning modality-specific representations with self-supervised multi-task learning for multimodal sentiment analysis.
\newblock In \emph{Thirty-Fifth {AAAI} Conference on Artificial Intelligence, {AAAI} 2021, Thirty-Third Conference on Innovative Applications of Artificial Intelligence, {IAAI} 2021, The Eleventh Symposium on Educational Advances in Artificial Intelligence, {EAAI} 2021, Virtual Event, February 2-9, 2021}, pages 10790--10797. {AAAI} Press.

\bibitem[{Zadeh et~al.(2017)Zadeh, Chen, Poria, Cambria, and Morency}]{emnlp:tfn}
Amir Zadeh, Minghai Chen, Soujanya Poria, Erik Cambria, and Louis-Philippe Morency. 2017.
\newblock \href {https://doi.org/10.18653/v1/D17-1115} {Tensor fusion network for multimodal sentiment analysis}.
\newblock In \emph{Proceedings of the 2017 Conference on Empirical Methods in Natural Language Processing}, pages 1103--1114, Copenhagen, Denmark. Association for Computational Linguistics.

\bibitem[{Zadeh et~al.(2018)Zadeh, Liang, Mazumder, Poria, Cambria, and Morency}]{aaai:mfn}
Amir Zadeh, Paul~Pu Liang, Navonil Mazumder, Soujanya Poria, Erik Cambria, and Louis{-}Philippe Morency. 2018.
\newblock Memory fusion network for multi-view sequential learning.
\newblock In \emph{Proceedings of the Thirty-Second {AAAI} Conference on Artificial Intelligence, (AAAI-18), the 30th innovative Applications of Artificial Intelligence (IAAI-18), and the 8th {AAAI} Symposium on Educational Advances in Artificial Intelligence (EAAI-18), New Orleans, Louisiana, USA, February 2-7, 2018}, pages 5634--5641. {AAAI} Press.

\bibitem[{Zadeh et~al.(2016)Zadeh, Zellers, Pincus, and Morency}]{arxiv:mosi}
Amir Zadeh, Rowan Zellers, Eli Pincus, and Louis{-}Philippe Morency. 2016.
\newblock {MOSI:} multimodal corpus of sentiment intensity and subjectivity analysis in online opinion videos.
\newblock \emph{arXiv preprint arXiv:1606.06259}.

\bibitem[{Zhang et~al.(2021)Zhang, Ju, Zhang, Li, Li, Zhu, and Zhou}]{aaai:hhmpn}
Dong Zhang, Xincheng Ju, Wei Zhang, Junhui Li, Shoushan Li, Qiaoming Zhu, and Guodong Zhou. 2021.
\newblock \href {https://doi.org/10.1609/AAAI.V35I16.17686} {Multi-modal multi-label emotion recognition with heterogeneous hierarchical message passing}.
\newblock In \emph{Thirty-Fifth {AAAI} Conference on Artificial Intelligence, {AAAI} 2021, Thirty-Third Conference on Innovative Applications of Artificial Intelligence, {IAAI} 2021, The Eleventh Symposium on Educational Advances in Artificial Intelligence, {EAAI} 2021, Virtual Event, February 2-9, 2021}, pages 14338--14346. {AAAI} Press.

\bibitem[{Zhang et~al.(2022)Zhang, Xu, Wang, Zhou, Zhao, and Teng}]{mm:mintrec}
Hanlei Zhang, Hua Xu, Xin Wang, Qianrui Zhou, Shaojie Zhao, and Jiayan Teng. 2022.
\newblock \href {https://doi.org/10.1145/3503161.3547906} {Mintrec: {A} new dataset for multimodal intent recognition}.
\newblock In \emph{{MM} '22: The 30th {ACM} International Conference on Multimedia, Lisboa, Portugal, October 10 - 14, 2022}, pages 1688--1697. {ACM}.

\bibitem[{Zhang et~al.(2023)Zhang, Wang, Yin, Liu, Liu, and Yu}]{emnlp:almt}
Haoyu Zhang, Yu~Wang, Guanghao Yin, Kejun Liu, Yuanyuan Liu, and Tianshu Yu. 2023.
\newblock \href {https://doi.org/10.18653/v1/2023.emnlp-main.49} {Learning language-guided adaptive hyper-modality representation for multimodal sentiment analysis}.
\newblock In \emph{Proceedings of the 2023 Conference on Empirical Methods in Natural Language Processing}, pages 756--767, Singapore. Association for Computational Linguistics.

\bibitem[{Zheng et~al.(2024)Zheng, Peng, Wang, Zhu, Guo, and Cai}]{eswa:fnenet}
Cangzhi Zheng, Junjie Peng, Lan Wang, Li’an Zhu, Jiatao Guo, and Zesu Cai. 2024.
\newblock Frame-level nonverbal feature enhancement based sentiment analysis.
\newblock \emph{Expert Systems with Applications}, 258:125148.

\bibitem[{Zong et~al.(2023)Zong, Ding, Li, Li, Zheng, and Zhou}]{mm:acformer}
Daoming Zong, Chaoyue Ding, Baoxiang Li, Jiakui Li, Ken Zheng, and Qunyan Zhou. 2023.
\newblock Acformer: An aligned and compact transformer for multimodal sentiment analysis.
\newblock In \emph{Proceedings of the 31st {ACM} International Conference on Multimedia, {MM} 2023, Ottawa, ON, Canada, 29 October 2023- 3 November 2023}, pages 833--842. {ACM}.

\end{thebibliography}

\appendix
\section{Formal Lemma and Proof}
\label{sec:Lemma and Proof}

Following previous research \citep{acl:mult, pr:tetfn}, we use only low-level temporal feature sequences $\mathbf{X}_{\{t,v,a\}}$ as the input for multimodal fusion. To better illustrate, $\mathbf{X}_{\{t,v,a\}}$ is considered as the graph vertex sequence (set) $\mathcal{V}_{\{t,v,a\}}$. These vertices are then concatenated into a single sequence $\mathcal{V}_m = [\mathcal{V}_t; \mathcal{V}_v; \mathcal{V}_a]^{\top}$. $\mathcal{V}_m$ is treated as the multimodal graph embedding (MGE), which is also regarded as the multimodal vertex set. We define $\mathcal{W}_u \in \mathbb{R}^{d_u \times d_u^f}$, where $u \in \{q, k, v\}$, $d_u$ is the original feature dimension of the vertices, $d_u^f$ is the attention feature dimension, as the projection weights for queries, keys, and values of $\mathcal{V}_m$ and $\mathcal{V}_m^b$. The information fusion operation from $\mathcal{V}_j$ to $\mathcal{V}_i$ is briefly defined as $j \rightarrow i$.

\textbf{Graph Structure Construction.} First, we use the self-attention mechanism as the fundamental theory to construct a block attention weight matrix $\mathcal{A}$ with modality combinations as units. In $\mathcal{A}$, $\mathcal{E}^{i,j} \in \mathbb{R}^{T_i \times T_j}$, where $\{i,j\} \in \{t, v, a\}$, is the attention weight submatrix constructed from $\mathcal{V}_i$ and $\mathcal{V}_j$ with $i$ as the query and $j$ as the key-value. It should be noted that the weight matrix has not yet been processed by the softmax function and cannot be used directly.

\begin{equation}
\begin{aligned}
\label{eq:adjmat_all}
  \mathcal{A} = (\mathcal{W}_q \mathcal{V}_m) \cdot (\mathcal{W}_k \mathcal{V}_m)^{\top} = \begin{pmatrix}
          \mathcal{E}^{t,t} & \mathcal{E}^{t,v} & \mathcal{E}^{t,a} \\
          \mathcal{E}^{v,t} & \mathcal{E}^{v,v} & \mathcal{E}^{v,a} \\
          \mathcal{E}^{a,t} & \mathcal{E}^{a,v} & \mathcal{E}^{a,a} \\
        \end{pmatrix}
\end{aligned}
\end{equation}

\textbf{Vertex Aggregation.} Assume a set of vertex features $\mathcal{V} = \{v_1, v_2, \dots, v_N\}$, where $v_i \in \mathbb{R}^D$, with $N$ being the number of vertices and $D$ the feature dimension of each vertex.

Based on previous work \citep{iclr:gat, iclr:gatv2}, the Graph Attention Network (GAT) is defined as follows. GAT performs self-attention over the vertices, which is a shared attention mechanism $a: \mathbb{R}^{D'} \times \mathbb{R}^D \rightarrow \mathbb{R}$ that computes attention coefficients. Before this, a shared linear transformation parameterized by a weight matrix $\mathbf{W} \in \mathbb{R}^{D' \times D}$ is applied.

\begin{equation}
   e^{i,j} = a(\mathbf{W}v_i, \mathbf{W}v_j) = (\mathbf{W}v_i) \cdot (\mathbf{W}v_j)^\top
\end{equation}

$e^{ij}$ represents the importance of vertex $j$'s features to vertex $i$. In the most general formulation, the model allows a vertex to attend to every other vertex, which discards all structural information. GAT injects the graph structure into the mechanism by performing masked attention: it only computes $e^{ij}$ for $j \in \mathcal{N}_i$, where $\mathcal{N}_i$ is the neighborhood of vertex $i$ in the graph. To make the coefficients comparable across different vertices, GAT normalizes them using the softmax function ($\mathcal{S}$):

\begin{equation}
  \alpha^{i,j} = \mathcal{S}_j(e^{i,j}) = \frac{\exp(e^{i,j})}{\sum_{k \in \mathcal{N}_i} \exp(e^{i,k})}
\end{equation}

Thus, the final output feature for each vertex is defined as follows:

\begin{equation}
  \overline{v}_i = \sum_{j \in \mathcal{N}_i} \alpha^{i,j} \mathbf{W}v_j
\end{equation}

Then, we extend the mechanism to multi-head attention. The concatenation operation for feature dimension tensors is denoted by $\parallel$. $L$ refers to the number of heads in the multi-head self-attention.

\begin{equation}
  \overline{v}_i = \parallel_{l=1}^L \sum_{j \in \mathcal{N}_i} \alpha^{i,j}_l \mathbf{W}_l v_j
\end{equation}

The above equations describe how to effectively aggregate vertex features by constructing the graph structure and applying the multi-head self-attention mechanism. 

\textbf{From Vertex Set to Vertex Aggregation.} Assume there are two vertex sets $\mathcal{V}_i = \{v^1_i, v^2_i, \dots, v^{N_i}_i\}$, where $v^m_i \in \mathbb{R}^{D_i}$, and $\mathcal{V}_j = \{v^1_j, v^2_j, \dots, v^{N_j}_j\}$, where $v^n_j \in \mathbb{R}^{D_j}$. Here, $N_{\{i,j\}}$ is the number of vertices in $\mathcal{V}_{\{i,j\}}$, and $D_{\{i,j\}}$ is the feature dimension of each vertex in $\mathcal{V}_{\{i,j\}}$.

Then, the GAT algorithm is applied to $v^m_i$ and $v^n_j$. However, instead of using a shared linear transformation, we use two independent weight matrices: the query weight $\mathbf{W}_q^m \in \mathbb{R}^{D_i' \times D_i}$ and the key weight $\mathbf{W}_k^n \in \mathbb{R}^{D_j' \times D_j}$. Let $\mathcal{N}_m$ denote the set of indices of vertices in $\mathcal{V}_j$ that are connected to $v^m_i$.

\begin{align}
    e^{m,n} &= a(\mathbf{W}_q^m v^m_i, \mathbf{W}_k^n v^n_j) \\
    \alpha^{m,n} &= \mathcal{S}_n(e^{m,n}) = \frac{\exp(e^{m,n})}{\sum_{l \in \mathcal{N}_m} \exp(e^{m,l})}
\end{align}

Next, we compute the final output feature for $v^m_i$. We apply the value weight $\mathbf{W}_v^m \in \mathbb{R}^{D_i' \times D_i}$ to transform $v^n_j$:

\begin{equation}
\label{eq:vtx_sbg_0}
  \overline{v}_m^i = \parallel_{l=1}^L \sum_{n \in \mathcal{N}_m} \alpha^{m,n}_l \mathbf{W}_{v_l}^n v^n_j
\end{equation}

Assuming $\mathcal{N}_m$ includes all vertices in the vertex set $\mathcal{V}_j$, i.e., $\mathcal{N}_m = N_j$, the current attention weight matrix is a vector $\mathcal{G}^m$. Since the two vertex sets $\{v_i^m\}$ and $\mathcal{V}_j$ are disjoint, $\mathcal{G}^m$ can be seen as the adjacency matrix of a unidirectional complete bipartite graph from the vertex set $\mathcal{V}_j$ to the vertex $v^m_i$, which preserves all edges and their weights between the two vertex sets. The bipartite graph is represented as follows:

\begin{equation}
    \textbf{G}_{i,j}^m = (\{v_i^m\}, \mathcal{V}_j, \mathcal{G}^m)
\end{equation}

The key and value weights for $\mathcal{V}_j$ are denoted as $\mathcal{W}_{\{k, v\}} \in \mathbb{R}^{N_j \times D_j' \times D_j}$. Subsequently, the aggregation process can be fully defined.

\begin{align}
\label{eq:vtx_sbg_1}
  e^m &= a(\mathbf{W}_q^m v_m^i, \mathcal{W}_k \mathcal{V}_j), \quad \mathcal{G}^m = \mathcal{S}(e^m) \\
\label{eq:vtx_sbg_2}
  \overline{v}_m^i &= \parallel_{l=1}^L (\mathcal{G}^m_l \mathcal{W}_{v_l} \mathcal{V}_j)
\end{align}

Here, the linear transformation weights for vertices are converted to the form of linear transformation weights for vertex sets by concatenating the tensors along the time series dimension.

\begin{align}
    \mathcal{W}_k &= [\mathbf{W}_k^1; \mathbf{W}_k^2; \dots; \mathbf{W}_k^{N_j}] \\
    \mathcal{W}_v &= [\mathbf{W}_v^1; \mathbf{W}_v^2; \dots; \mathbf{W}_v^{N_j}]
\end{align}





\textbf{From Set to Set Aggregation.} Apply the algorithm defined by Equations \ref{eq:vtx_sbg_0}, \ref{eq:vtx_sbg_1}, and \ref{eq:vtx_sbg_2} to all vertices in $\mathcal{V}_i$. The aggregation form is now from a vertex set to another vertex set, thus, we transform the vertex-to-vertex aggregation into a unidirectional complete bipartite graph aggregation. The form also changes from scalar attention weights $e$ to an attention weight matrix $\mathcal{E}$, which we represent as the adjacency matrix of a unidirectional complete bipartite graph. The query weights for $\mathcal{V}_i$ are denoted as $\mathcal{W}_q$.

\begin{align}
\mathcal{W}_q &= [\mathbf{W}_q^1; \mathbf{W}_q^2; \dots; \mathbf{W}_q^{N_i}] \\
\label{eq:sbg_sbg0}
  \mathcal{E}^{i,j} &= a(\mathcal{W}_q \mathcal{V}_i, \mathcal{W}_k \mathcal{V}_j), \quad \mathcal{G}^{i, j} = \mathcal{S}(\mathcal{E}^{i,j}) \\
\label{eq:sbg_sbg1}
   \mathcal{V}^{'}_i &= \parallel_{l=1}^L (\mathcal{G}_l^{i,j} \mathcal{W}_{v_l} \mathcal{V}_j)
\end{align}

The aggregation process is now equivalent to the multi-head cross-modal attention mechanism in MulTs. Similarly, it is also equivalent to the multi-head cross-attention mechanism in the traditional Transformer decoder \citep{nips:transformer}. Therefore, we introduce the following lemma.

\begin{lemma}
    The multi-head cross-modal attention mechanism is equivalent to the aggregation of unidirectional complete bipartite graphs of bi-modality combination; the multi-head self-attention mechanism is equivalent to the aggregation of directed complete graphs of uni-modalities.
\end{lemma}

The proof of the lemma is straightforward:

(i) If $i \neq j$, the two vertex sets $\mathcal{V}_i$ and $\mathcal{V}_j$ are disjoint, and $\mathcal{G}^{i, j}$, together with the two sets, forms a unidirectional complete bipartite graph, with the direction from $j$ to $i$.

\begin{equation}
    \label{subgraph_def}
    \textbf{G}_{i,j} = (\mathcal{V}_i, \mathcal{V}_j, \mathcal{G}^{i, j})
\end{equation}

Completing the aggregation of $\textbf{G}_{i,j}$ is equivalent to performing the multi-head cross-modal attention calculation between $\mathcal{V}_i$ and $\mathcal{V}_j$ with $\mathcal{V}_i$ as the dominant modality.

(ii) If $i = j$, then $\mathcal{V}_i$ and $\mathcal{V}_j$ are the same set and are not disjoint. In this case, there is actually only one set, which we can denote as $\mathcal{V}_i$, with the adjacency matrix $\mathcal{G}^{i,i}$. The operation here actually conforms to the multi-head self-attention mechanism, forming a directed complete graph.

\begin{equation}
    \textbf{G}_{i,i} = (\mathcal{V}_i, \mathcal{G}^{i,i})
\end{equation}

Completing the aggregation of $\textbf{G}_{i,i}$ is equivalent to performing the multi-head self-attention calculation on $\mathcal{V}_i$.

\section{Computational and Space Complexity}
\label{complexity}

Assuming number of layers as $L \in \text{R}$, batch size as $\mathcal{B} \in \text{R}$, and $MLP$ weights as $\mathcal{W}_1 \in \mathbb{R}^{d_u^f \times d_u^p}, \mathcal{W}_2 \in \mathbb{R}^{d_u^p \times d_u^f}$, where $u\in\{m,t,v,a\}$. In particular, $d_u^f \leq d_u^p$, $d_m^f = d_{\{t,v,a\}}^f$, and $d_m^p = d_{\{t,v,a\}}^p$. We define computational complexity as $\textbf{C}_u^i$, space complexity as $\textbf{S}_u^i$, where $u \in \{mult, gsit\}$, $i$ denotes the function step.

\subsection{Computational Complexity}

\subsubsection{MulT}

In this section, we discuss the computational complexity of MulT \citep{acl:mult}. 

\textbf{QKV Projection 1}: assuming $i\in\{t,v,a\}, j\in\{t,v,a\} \setminus \{i\}$. The computational complexity $\textbf{C}_{mult}^{qkv_1}$ is:

\begin{equation}
    \begin{aligned}
     \textbf{C}_{mult}^{qkv_1} & = O(\sum(T_i {d_m^f}^2 + 2T_j {d_m^f}^2)) \\
        & = O(3(T_t + T_v + T_a){d_m^f}^2 \times 2) \\
        & = O((T_t + T_v + T_a)6{d_m^f}^2)
    \end{aligned}
\end{equation}

\textbf{Attention 1}: assuming $i\in\{t,v,a\}, j\in\{t,v,a\} \setminus \{i\}$.

First, generate attention maps, then apply scaling and softmax function. The computational complexity $\textbf{C}_{mult}^{attn_1^1}$ is:

\begin{equation}
    \begin{aligned}
     \textbf{C}_{mult}^{attn_1^1}  & = O(\sum (T_i T_j d_m^f + 2T_i T_j)) \\
     & =  O((T_t T_v + T_t T_a + T_v T_a)(2d_m^f + 4))
    \end{aligned}
\end{equation}

Then, perform aggregation. $\textbf{C}_{mult}^{attn_1^1}$ is as follows:

\begin{equation}
    \begin{aligned}
      \textbf{C}_{mult}^{attn_1^2}  
      & = O(\sum (T_i T_j d_m^f)) \\
      & = O((T_t T_v + T_t T_a + T_v T_a)2d_m^f)
    \end{aligned}
\end{equation}

Thus, the overall complexity is clear.

\begin{equation}
    \begin{aligned}
    \textbf{C}_{mult}^{attn_1} &= \textbf{C}_{mult}^{attn_1^1} + \textbf{C}_{mult}^{attn_1^2} \\ 
       &= O((T_t T_v + T_t T_a + T_v T_a)(4d_m^f + 4))
    \end{aligned}
\end{equation}

\textbf{MLP 1}: assuming $i\in\{t,v,a\}$. The computational complexity $\textbf{C}_{mult}^{mlp_1}$ is:

\begin{equation}
    \begin{aligned}
       \textbf{C}_{mult}^{mlp_1} &= O(\sum 2(T_i d_m^f d_m^p)) \\
        &= O((T_t + T_v + T_a)2 d_m^f d_m^p)
    \end{aligned}
\end{equation}

\textbf{QKV Projection 2}: assuming $i\in\{t,v,a\}$. The computational complexity $\textbf{C}_{mult}^{qkv_2}$ is: 

\begin{equation}
    \begin{aligned}
       \textbf{C}_{mult}^{qkv_2} &= O(\sum(3T_i (2d_m^f)^2)) \\
        &= O((T_t + T_v + T_a)12{d_m^f}^2) 
    \end{aligned}
\end{equation}

\textbf{Attention 2}: assuming $i\in\{t,v,a\}$. 

First, generate attention maps, then apply scaling and softmax function. The computational complexity $\textbf{C}_{mult}^{attn_2^1}$ is:

\begin{equation}
    \begin{aligned}
       \textbf{C}_{mult}^{attn_2^1} &=  O(\sum (T_i^2 2d_m^f + 2T_i^2)) \\
        &= O((T_t^2 + T_v^2 + T_a^2)(4d_m^f + 4))
    \end{aligned}
\end{equation}

Then, perform aggregation. The computational complexity $\textbf{C}_{mult}^{attn_2^2}$ is:

\begin{equation}
    \begin{aligned}
        \textbf{C}_{mult}^{attn_2^2} &= O(\sum (T_i^2 2d_m^f)) \\
        &=  O((T_t^2 + T_v^2 + T_a^2)2d_m^f)
    \end{aligned}
\end{equation}

Thus, the overall complexity $\textbf{C}_{mult}^{attn_2}$ is clear.

\begin{equation}
    \begin{aligned}
    \textbf{C}_{mult}^{attn_2} &= \textbf{C}_{mult}^{attn_2^1} + \textbf{C}_{mult}^{attn_2^2} \\ 
    &= O((T_t^2 + T_v^2 + T_a^2)(6d_m^f + 4))
    \end{aligned}
\end{equation}

\textbf{MLP 2}: assuming $i\in\{t,v,a\}$. The computational complexity $\textbf{C}_{mult}^{mlp_2}$ is:

\begin{equation}
    \begin{aligned}
       \textbf{C}_{mult}^{mlp_2} &= O(\sum 2(T_i 2d_m^f 2d_m^p)) \\
        &= O((T_t + T_v + T_a)8d_m^f d_m^p)
    \end{aligned}
\end{equation}

\subsubsection{GsiT}

In this section, we discuss the computational complexity of GsiT. Note that $T_m = T_t + T_v + T_a$.

\textbf{QKV Projection 1}: assuming $i\in\{t,v,a\}, j\in\{t,v,a\} \setminus \{i\}$. The computational complexity $\textbf{C}_{gsit}^{qkv_1}$ is:

\begin{equation}
    \begin{aligned}
        \textbf{C}_{gsit}^{qkv_1} &= O(3T_m{d_m^f}^2 \times 3) \\
        &= O((T_t + T_v + T_a)6{d_m^f}^2)
    \end{aligned}
\end{equation}

\textbf{Attention 1}: assuming $i\in\{t,v,a\}$.

First, generate attention maps, then apply scaling and softmax function. If we explicitly add the mask, it will be exceedingly high in complexity. The computational complexity $\textbf{C}_{gsit}^{attn_1^1}$ is:

\textit{w / o Decomposition}:

\begin{equation}
    \begin{aligned}
        \textbf{C}_{gsit}^{attn_1^1} &= O(2T_m^2 (d_m^f + 3)) \\
        &= O((T_t + T_v + T_a)^2(2d_m^f + 6))
    \end{aligned}
\end{equation}

However, if we decompose the multimodal sequences inside of the procedure, it will be.

\textit{w /  Decomposition}:

\begin{equation}
    \begin{aligned}
        \textbf{C}_{gsit}^{attn_1^1} &= O(\sum (T_i T_j d_m^f + 2T_i T_j)) \\
        &= O((T_t T_v + T_t T_a + T_v T_a)(2d_m^f + 4))
    \end{aligned}
\end{equation}

Then, perform aggregation. The computational complexity $\textbf{C}_{gsit}^{attn_1^2}$ is:

\textit{w / o Decomposition}:

\begin{equation}
    \begin{aligned}
        \textbf{C}_{gsit}^{attn_1^2} &= O(2T_m^2 d_m^f - 2\times\sum(T_i^2 d_m^f)) \\
        &= O((T_t T_v + T_t T_a + T_v T_a)2d_m^f)
    \end{aligned}
\end{equation}

\textit{w / Decomposition}:

\begin{equation}
    \begin{aligned}
        \textbf{C}_{gsit}^{attn_1^2} &= O(\sum (T_i T_j d_m^f)) \\
        &= O((T_t T_v + T_t T_a + T_v T_a)2d_m^f)
    \end{aligned}
\end{equation}

Thus, the overall complexity $\textbf{C}_{gsit}^{attn_1}$ is clear.

\textit{w/o Decomposition}:

\begin{equation}
    \begin{aligned}
        \textbf{C}_{gsit}^{attn_1} &= \textbf{C}_{gsit}^{attn_1^2} + \textbf{C}_{gsit}^{attn_1^2} \\
        &= O((T_t + T_v + T_a)^2(2d_m^f + 6) \\
        &+ (T_t T_v + T_t T_a + T_v T_a)2d_m^f)
    \end{aligned}
\end{equation}

\textit{w/ Decomposition}:

\begin{equation}
    \begin{aligned}
    \textbf{C}_{gsit}^{attn_1} &= \textbf{C}_{gsit}^{attn_1^2} + \textbf{C}_{gsit}^{attn_1^2} \\
    &= O((T_t T_v + T_t T_a + T_v T_a)(4d_m^f + 4))
    \end{aligned}
\end{equation}

\textbf{MLP 1}: apply $MLP$ to $T_m$. The computational complexity $\textbf{C}_{gsit}^{mlp_1}$ is:

\begin{equation}
    \begin{aligned}
       \textbf{C}_{gsit}^{mlp_1} &= O(2(T_m d_m^f d_m^p)) \\
        &= O((T_t + T_v + T_a) 2 d_m^f d_m^p)
    \end{aligned}
\end{equation}

\textbf{QKV Projection 2}: The computational complexity $\textbf{C}_{gsit}^{qkv_2}$ is:

\begin{equation}
    \begin{aligned}
        \textbf{C}_{gsit}^{qkv_2} &= O(3T_m (2d_m^f)^2)) \\
        &= O((T_t + T_v + T_a)12{d_m^f}^2) 
    \end{aligned}
\end{equation}

\textbf{Attention 2}: assuming $i\in\{t,v,a\}, j\in\{t,v,a\} \setminus \{i\}$.

First, generate attention maps, then apply scaling and softmax function. The computational complexity $\textbf{C}_{gsit}^{attn_2^1}$ is:

\textit{w / o Decomposition}:

\begin{equation}
    \begin{aligned}
        \textbf{C}_{gsit}^{attn_2^1} &= O({T_m}^2 (2d_m^f + 3)) \\
        &= O((T_t + T_v + T_a)^2(2d_m^f + 3))
    \end{aligned}
\end{equation}

\textit{w / Decomposition}:

\begin{equation}
    \begin{aligned}
        \textbf{C}_{gsit}^{attn_2^1} &= O(\sum ({T_i}^2 2d_m^f + 2{T_i}^2)) \\
        &= O(({T_t}^2 + {T_v}^2 + {T_a}^2)(2d_m^f + 2))
    \end{aligned}
\end{equation}

Then, perform aggregation.

\textit{w / o Decomposition}:

\begin{equation}
    \begin{aligned}
        \textbf{C}_{gsit}^{attn_2^2} &= O({T_m}^2 2 d_m^f - \sum (T_i T_j 2 d_m^f)) \\
        &= O(({T_t}^2 + {T_v}^2 + {T_a}^2)2d_m^f)
    \end{aligned}
\end{equation}

\textit{w / Decomposition}:

\begin{equation}
    \begin{aligned}
        \textbf{C}_{gsit}^{attn_2^2} &= O(\sum ({T_i}^2 2d_m^f)) \\
        &= O(({T_t}^2 + {T_v}^2 + {T_a}^2)2d_m^f)
    \end{aligned}
\end{equation}

Thus, the overall complexity is clear.

\textit{w / o Decomposition}:

\begin{equation}
    \begin{aligned}
    \textbf{C}_{gsit}^{attn_2} &= \textbf{C}_{gsit}^{attn_2^1} + \textbf{C}_{gsit}^{attn_2^2} \\
        &= O((T_t + T_v + T_a)^2(2d_m^f + 3)\\ 
        &+ ({T_t}^2 + {T_v}^2 + {T_a}^2)2d_m^f)
    \end{aligned}
\end{equation}

\textit{w / Decomposition}:

\begin{equation}
    \begin{aligned}
    \textbf{C}_{gsit}^{attn_2} &= \textbf{C}_{gsit}^{attn_2^1} + \textbf{C}_{gsit}^{attn_2^2} \\
       &= O(({T_t}^2 + {T_v}^2 + {T_a}^2)(4d_m^f + 2)
    \end{aligned}
\end{equation}

\textbf{MLP 2}: assuming $i\in\{t,v,a\}$. The computational complexity $\textbf{C}_{gsit}^{mlp_2}$ is:

\begin{equation}
    \begin{aligned}
       \textbf{C}_{gsit}^{mlp_2} &= O(2(T_m 2d_m^f 2d_m^p)) \\
        &= O((T_t + T_v + T_a)8d_m^f d_m^p)
    \end{aligned}
\end{equation}

\subsubsection{Overall Assessment}

In the decomposed pattern, the computational complexity of GsiT is equal to that of MulT.

\begin{equation}
    \begin{aligned}
       \Delta\textbf{C} \equiv O(0)
    \end{aligned}
\end{equation}

Without the decomposition, the computational complexity of GsiT exceeds as follows. We ignore the equal values.

\begin{equation}
    \begin{aligned}
       \Delta\textbf{C} &= \textbf{C}_{gsit}^{attn_1^1} + \textbf{C}_{gsit}^{attn_2^1} - (\textbf{C}_{mult}^{attn_1^1} + \textbf{C}_{mult}^{attn_2^1}) \\
        &= O(5(T_t + T_v + T_a)^2 \\
        &+ (T_t T_v + T_t T_a + T_v T_a) (6d + 4))
    \end{aligned}
\end{equation}

\subsection{Space Complexity}

\subsubsection{MulT}

In this section, we discuss the space complexity of MulT. We split it into two kinds: \textit{Parameter} and \textit{Runtime}.

\textbf{QKV Projection 1} (\textit{Parameter}): The space complexity $\textbf{S}_{mult}^{qkv_1}$ is:

\begin{equation}
   \textbf{S}_{mult}^{qkv_1} = O(3{d_m^f}^2 \times 3) = O(9{d_m^f}^2)
\end{equation}

\textbf{Attention 1} (\textit{Runtime}): assuming $i\in\{t,v,a\}, j\in\{t,v,a\} \setminus \{i\}$. Due to the decoupling in the computational process of different bi-modality combinations, each combination's attention map independently occupies GPU memory during its computation and is released upon completion. The space complexity $\textbf{S}_{mult}^{attn_1}$ is: 

\begin{equation}
   \textbf{S}_{mult}^{attn_1} = O(T_iT_j)
\end{equation}

\textbf{MLP 1} (\textit{Parameter}): The space complexity $\textbf{S}_{mult}^{mlp_1}$ is:

\begin{equation}
    \textbf{S}_{mult}^{mlp_1} = O(2d_m^fd_m^p \times 3) = O(6d_m^fd_m^p)
\end{equation}

\textbf{QKV Projection 2} (\textit{Parameter}): The space complexity $\textbf{S}_{mult}^{qkv_2}$ is:

\begin{equation}
    \textbf{S}_{mult}^{qkv_2} = O(3\times{2d_m^f}^2 \times 3) = O(36{d_m^f}^2)
\end{equation}

\textbf{Attention 2} (\textit{Runtime}): assuming $i\in\{t,v,a\}$. The space complexity $\textbf{S}_{mult}^{attn_2}$ is:

\begin{equation}
    \textbf{S}_{mult}^{attn_2} = O({T_i}^2)
\end{equation}

\textbf{MLP 2} (\textit{Parameter}): The space complexity $\textbf{S}_{mult}^{mlp_2}$ is:

\begin{equation}
    \textbf{S}_{mult}^{mlp_2} = O(2\times2d_m^f2d_m^p \times 3) = O(24d_m^fd_m^p)
\end{equation}

\subsubsection{GsiT}

In this section, we discuss the space complexity of GsiT.

\textbf{QKV Projection 1} (\textit{Parameter}): shared weights. The space complexity $\textbf{S}_{gsit}^{qkv_1}$ is:

\begin{equation}
    \textbf{S}_{gsit}^{qkv_1} = O(3{d_m^f}^2) = O(3{d_m^f}^2)
\end{equation}

\textbf{Attention 1} (\textit{Runtime}): assuming $i\in\{t,v,a\}, j\in\{t,v,a\} \setminus \{i\}$. The space complexity $\textbf{S}_{gsit}^{attn_1}$ is:

\textit{w / o Decomposition}:

\begin{equation}
    \textbf{S}_{gsit}^{attn_1} = O(T_m^2) = O((T_t + T_v + T_a)^2)
\end{equation}

\textit{w / Decomposition}:

\begin{equation}
    \textbf{S}_{gsit}^{attn_1} = O(T_iT_j)
\end{equation}

\textbf{MLP 1} (\textit{Parameter}): The space complexity $\textbf{S}_{gsit}^{mlp_1}$ is:

\begin{equation}
    \textbf{S}_{gsit}^{mlp_1} = O(2d_m^fd_m^p)
\end{equation}

\textbf{QKV Projection 2} (\textit{Parameter}): The space complexity $\textbf{S}_{gsit}^{qkv_2}$ is:

\begin{equation}
    \textbf{S}_{gsit}^{qkv_2} = O(3\times({2d_m^f})^2) = O(12{d_m^f}^2)
\end{equation}

\textbf{Attention 2} (\textit{Runtime}): assuming $i\in\{t,v,a\}$. The space complexity $\textbf{S}_{gsit}^{attn_2}$ is:

\textit{w / o Decomposition}:

\begin{equation}
    \textbf{S}_{gsit}^{attn_2} = O({T_m}^2) = O((T_t + T_v + T_a)^2)
\end{equation}

\textit{w / Decomposition}:

\begin{equation}
    \textbf{S}_{gsit}^{attn_2} = O({T_i}^2)
\end{equation}

\textbf{MLP 2} (\textit{Parameter}): The space complexity $\textbf{S}_{gsit}^{mlp_2}$ is:

\begin{equation}
    \textbf{S}_{gsit}^{mlp_2} = O(2\times2d_m^f2d_m^p) = O(8d_m^fd_m^p)
\end{equation}

\subsubsection{Overall Assessment}

GsiT has 2/3 fewer static parameters compared to MulT. When using only the Interlaced Mask, GsiT's runtime GPU memory usage is significantly higher than that of MulT. However, by applying the Decomposition operation, the GPU memory usage can be reduced to the same level as MulT. Specifically, take $F \in \{qkv_1, mlp_1, qkv_2, mlp_2\}$, it turns out to be:

\textbf{Parameter}: 

\begin{equation}
    \begin{aligned}
        &\Delta \textbf{S}_1 = \sum_{u \in F} \textbf{S}_{mult}^u - \sum_{u \in F}\textbf{S}_{gsit}^u \\ 
        &\frac{\Delta \textbf{S}_1}{\sum_{u \in F}\textbf{S}_{mult}^u} =  \frac{1}{3}
    \end{aligned}
\end{equation}

\textbf{Runtime}:

\textit{(i) w/o Decomposition}: assuming $T_a > T_v > T_t$. Runtime space complexity is dynamic and we need to compare step by step.

\begin{equation}
    \begin{aligned}
       \textbf{S}_{gsit}^{attn_2} - O({T_a}^2) \leq &\Delta \textbf{S}_2 \leq \textbf{S}_{gsit}^{attn_1} - O{T_tT_v} \\ 
       O({T_m}^2 - {T_a}^2) \leq &\Delta \textbf{S}_2 \leq O({T_m}^2 - T_tT_a)
    \end{aligned}
\end{equation}

\textit{(ii) w/ Decomposition}: 
\begin{equation}
    \begin{aligned}
        \Delta \textbf{S}_2 \equiv O(0)
    \end{aligned}
\end{equation}

\section{Graph Structures}
\label{graph_structure}

\textit{Original Structure}: The original structure is defined as two opposing unidirectional ring graphs. They both realize cyclic all-modal-in-one fusion, which makes trimodal information fully interact in shared model weights. The structure is: $\{t \rightarrow v, v \rightarrow a, a \rightarrow t\}$, $\{a \rightarrow v, v \rightarrow t, t \rightarrow a \}$. The modal-wise IFMs are:

\vspace{-0.3cm} 
\begin{equation}
\begin{aligned}
    \mathcal{M}_{inter}^{forward} = \begin{pmatrix}
        \mathcal{J}^{t,t} & \mathcal{O}^{t,v} & \mathcal{J}^{t,a} \\
        \mathcal{J}^{v,t} & \mathcal{J}^{v,v} & \mathcal{O}^{v,a} \\
        \mathcal{O}^{a,t} & \mathcal{J}^{a,v} & \mathcal{J}^{a,a} \\
      \end{pmatrix} \\ \mathcal{M}_{inter}^{backward} = \begin{pmatrix}
        \mathcal{J}^{t,t} & \mathcal{J}^{v,t} & \mathcal{O}^{a,t} \\
        \mathcal{O}^{v,t} & \mathcal{J}^{v,v} & \mathcal{J}^{v,a} \\
        \mathcal{J}^{a,t} & \mathcal{O}^{a,v} & \mathcal{J}^{a,a} \\
      \end{pmatrix}
\end{aligned}
\end{equation}

\textit{Structure-1}: Structure-1 realizes all-modal-in-one fusion, but the information passing is not cyclic. The structure is: $\{a \rightarrow v, v \rightarrow a, a \rightarrow t\}$, $\{v \rightarrow t, t \rightarrow v, t \rightarrow a\}$. The modal-wise IFMs are:

\vspace{-0.3cm} 
\begin{equation}
\begin{aligned}
    \mathcal{M}_{inter}^{forward} = \begin{pmatrix}
        \mathcal{J}^{t,t} & \mathcal{J}^{t,v} & \mathcal{O}^{t,a} \\
        \mathcal{J}^{v,t} & \mathcal{J}^{v,v} & \mathcal{O}^{v,a} \\
        \mathcal{O}^{a,t} & \mathcal{J}^{a,v} & \mathcal{J}^{a,a} \\
      \end{pmatrix}\\ \mathcal{M}_{inter}^{backward} = \begin{pmatrix}
        \mathcal{J}^{t,t} & \mathcal{O}^{v,t} & \mathcal{J}^{t,a} \\
        \mathcal{O}^{v,t} & \mathcal{J}^{v,v} & \mathcal{J}^{v,a} \\
        \mathcal{O}^{a,t} & \mathcal{J}^{a,v} & \mathcal{J}^{a,a} \\
      \end{pmatrix}
\end{aligned}
\end{equation}

\textit{Structure-2}: Structure-2 realizes all-modal-in-one fusion, but the information passing is not cyclic. The structure is: $\{v \rightarrow t, t \rightarrow v, v \rightarrow a\}$, $\{a \rightarrow t, t \rightarrow a, a \rightarrow v\}$. The modal-wise IFMs are:

\vspace{-0.3cm} 
\begin{equation}
\begin{aligned}
    \mathcal{M}_{inter}^{forward} = \begin{pmatrix}
        \mathcal{J}^{t,t} & \mathcal{O}^{t,v} & \mathcal{J}^{t,a} \\
        \mathcal{O}^{v,t} & \mathcal{J}^{v,v} & \mathcal{J}^{v,a} \\
        \mathcal{J}^{a,t} & \mathcal{O}^{a,v} & \mathcal{J}^{a,a} \\
      \end{pmatrix} \\ \mathcal{M}_{inter}^{backward} = \begin{pmatrix}
        \mathcal{J}^{t,t} & \mathcal{J}^{v,t} & \mathcal{O}^{t,a} \\
        \mathcal{J}^{v,t} & \mathcal{J}^{v,v} & \mathcal{O}^{v,a} \\
        \mathcal{O}^{a,t} & \mathcal{J}^{a,v} & \mathcal{J}^{a,a} \\
      \end{pmatrix}
\end{aligned}
\end{equation}

\textit{Structure-3}: Structure-3 realizes all-modal-in-one fusion, but the information passing is not cyclic. The structure is: $\{a \rightarrow v, v \rightarrow a, v\rightarrow t\}$, $\{a \rightarrow t, t \rightarrow a, t \rightarrow v\}$. The modal-wise IFMs are:

\vspace{-0.3cm} 
\begin{equation}
\begin{aligned}
    \label{structure3}
    \mathcal{M}_{inter}^{forward} = \begin{pmatrix}
        \mathcal{J}^{t,t} & \mathcal{O}^{t,v} & \mathcal{J}^{t,a} \\
        \mathcal{J}^{v,t} & \mathcal{J}^{v,v} & \mathcal{O}^{v,a} \\
        \mathcal{J}^{a,t} & \mathcal{O}^{a,v} & \mathcal{J}^{a,a} \\
      \end{pmatrix} \\ \mathcal{M}_{inter}^{backward} = \begin{pmatrix}
        \mathcal{J}^{t,t} & \mathcal{J}^{v,t} & \mathcal{O}^{t,a} \\
        \mathcal{O}^{v,t} & \mathcal{J}^{v,v} & \mathcal{J}^{v,a} \\
        \mathcal{O}^{a,t} & \mathcal{J}^{a,v} & \mathcal{J}^{a,a} \\
      \end{pmatrix}
\end{aligned}
\end{equation}


\textit{Self-Only}: Self-Only mask only contains masks intra-modal subgraphs. The structure is $\{t \rightarrow t, v \rightarrow v, a \rightarrow a\}$. The modal-wise IFM is:

\vspace{-0.3cm} 
\begin{equation}
  \mathcal{M}_{inter} = \begin{pmatrix}
    \mathcal{J}^{t,t} & \mathcal{O}^{t,v} & \mathcal{O}^{t,a} \\
    \mathcal{O}^{v,t} & \mathcal{J}^{v,v} & \mathcal{O}^{v,a} \\
    \mathcal{O}^{a,t} & \mathcal{O}^{a,v} & \mathcal{J}^{a,a} \\
  \end{pmatrix}
\end{equation}

\section{Experimental Settings}
\label{experiment_settings}

All experiments are based on BERT \citep{naacl:bert}, and we use the most basic version, bert-base-uncased, which is used as the text modality encoder. Following previous works  \citep{eswa:fmlmsn, taffc:mtmd}, the feature extraction tools of different modalities in each dataset. BERT \citep{naacl:bert} for text, OpenFace \citep{wacv:openface}, OpenFace 2.0 \citep{fg:openface2}, and ResNet50 \citep{cvpr:resnet} for vision, COVAREP \citep{icassp:covarep}, LibROSA, and Wav2Vec2 \citep{neurips:wav2vec2} for audio. For each datasets, the extractors are shown in Table \ref{tab: extractor}.

\begin{table*}[htpb]
  \caption{The extractors of the main experiment.}
  \label{tab: extractor}
  \centering
  \resizebox{\linewidth}{!}  
  {
    \begin{tabular}{lccccc}
      \toprule
      Modal & CMU-MOSI & CMU-MOSEI & CH-SIMS & MIntRec \\
      \midrule
      Text   & bert-base-uncased & bert-base-uncased & bert-base-chinese & bert-base-uncased\\
      Vision & OpenFace & OpenFace & OpenFace2.0 & ResNet50\\
      Audio  & COVAREP & COVAREP & LibROSA & Wav2Vec2 \\
      \bottomrule
    \end{tabular}
  }
\end{table*}

The reported results are the average of multiple runs with 5 random seeds to ensure the reliability and stability of our findings.

All experiments are performed on the platform equipped with the following computing infrastructures: GPU: Nvidia GeForce RTX 3060 12G; CPU: AMD Ryzen 9 5900X 12-Core Processor.

\section{Datasets}
\label{datasets}

\begin{table*}[htpb]
    \centering
  \caption{Dataset basic information, sample distribution statistics, and data forms for MSA and MIR datasets. Note: for part Sample Distribution Statistics, data is in format negative (< 0)/neutral (= 0)/positive (> 0) sentiment intensity. Specifically, for MIntRec, data is in format express emotions / attitudes achieve goals For part Data Forms, data is in format text / vision / audio.}
  \label{tab: dataset_inf}
  \resizebox{\textwidth}{!}  
  {
    \begin{tabular}{lccccc}
      \toprule
      Description & CMU-MOSI & CMU-MOSEI & CH-SIMS & MIntRec \\
      \midrule
      \multicolumn{5}{c}{Basic Information} \\
      \midrule
      Language & English & English & Chinese & English \\
      Unimodal Labels & None & None & T,V,A &  None \\
      \midrule
      \multicolumn{5}{c}{Sample Distribution Statistics} \\
      \midrule
      Train & 552/53/679 & 4,738/3,540/8,084 & 742/207/419 & 749/585\\
      Validation & 92/13/124 & 506/433/932 &  248/69/139 & 249/196\\
      Test & 379/30/227 &  1,350/1,025/2,284 & 248/69/140 & 248/197 \\
      Total & 2,199 & 22,856 & 2,281 & 2,224\\
      \midrule
      \multicolumn{5}{c}{Data Forms} \\
      \midrule
      Sequence Length(Max) & 50/375/500 & 50/500/500 & 39/400/55 & 30/230/480 \\
      \midrule
      Average Length(Train) & 14/42/38 & 24/94/149 & 17/22/158 & 12/53/116 \\
      Average Length(Validation) & 14/43/37 & 25/100/156 & 17/21/154 & 12/56/121 \\
      Average Length(Test) & 16/52/49 & 25/95/153 & 17/21/157 & 13/56/122 \\
      \midrule
      Length Variance(Train) & 66/927/805 & 148/5,115/8,105 & 53/116/6,050 & 20/562/2,420 \\
      Length Variance(Validation) & 63/983/658 & 145/4,626/7,401 & 48/101/5,358 & 21/687/2,967 \\
      Length Variance(Test) & 91/1,773/1,526 & 141/5,254/8,325 & 51/108/5,647 & 24/727/3,175 \\
      \midrule
      Feature Dimension & 768/20/5 & 768/35/74 & 768/709/33 & 768/256/768\\
      \bottomrule
    \end{tabular}
  }
\end{table*}

Table \ref{tab: dataset_inf} shows a brief introduction to the chosen datasets. The detailed descriptions are as follows.

\textbf{CMU-MOSI} \citep{arxiv:mosi}: The CMU-MOSI is a widely used dataset for human multimodal sentiment analysis, containing 2,198 short monologue video clips. Each clip is a single-sentence utterance expressing the speaker's opinion on a topic like movies. The utterances are manually annotated with a continuous opinion score ranging from -3 to +3, where -3 represents highly negative, -2 negative, -1 weakly negative, 0 neutral, +1 weakly positive, +2 positive, and +3 highly positive.

\textbf{CMU-MOSEI} \citep{acl:mosei}: CMU-MOSEI is an improved version of CMU-MOSI, containing 23,453 annotated video clips (approximately 10 times more than CMU-MOSI) from 5,000 videos, involving 1,000 different speakers and 250 distinct topics. The dataset also features a larger number of discourses, samples, speakers, and topics compared to CMU-MOSI. The range of labels for each discourse remains consistent with CMU-MOSI.

\textbf{CH-SIMS} \citep{acl:ch-sims}: The CH-SIMS dataset includes the same modalities as CMU-MOSI: audio, text, and video, collected from 2281 annotated video segments. It features data from TV shows and movies, making it culturally distinct and diverse. Additionally, CH-SIMS provides multiple labels for the same utterance based on different modalities, adding an extra layer of complexity and richness to the data.

\textbf{MIntRec} \citep{mm:mintrec}: The MIntRec dataset  is a fine-grained dataset for multimodal intent recognition with 2,224 high-quality samples with text, video and audio modalities across 20 intent categories.

\section{Baselines}

\label{baselines}

\textbf{MulT} \citep{acl:mult}: Multimodal Transformer (MulT) achieves cross-modal translation using a cross-modal Transformer based on cross-modal attention. It was the first to propose the comprehensive fusion paradigm defined by Equation \ref{g_1}.

\textbf{Self-MM} \citep{aaai:self-mm}: Learning Modal-Specific Representations with Self-Supervised Multi-Task Learning (Self-MM) designs a multi- and a uni- task to learn inter-modal consistency and intra-modal specificity, being one of the most widely used representation learning frameworks in the MSA domain.

\textbf{TETFN} \citep{pr:tetfn}: Text Enhanced Transformer Fusion Network (TETFN) strengthens the role of text modes in multimodal information fusion through text-oriented cross-modal mapping and single-modal label generation, and uses Vision-Transformer pre-training model to extract visual features.

\textbf{ALMT} \citep{emnlp:almt}: The Adaptive Language-guided Multimodal Transformer (ALMT) incorporates an Adaptive Hyper-modality Learning (AHL) module to learn an unrelated or conflict-suppressing representation from visual and audio features under the guidance of language features at different scales.

\textbf{MMIM} \citep{emnlp:mmim}: MultiModal InfoMax (MMIM) hierarchically maximizes mutual information within unimodal features and between multimodal fusion features and unimodal features to obtain emotion-related information.

\section{Related Work}

\label{relatedworks}

Multimodal Sentiment Analysis (MSA) is an increasingly popular research area. The data form of MSA typically consists of two or more modalities, with the most widely used form being a tri-modality combination of text, visual, and audio. Multimodal fusion is the core issue in the MSA field, and early models mostly focus on it. 

\subsection{Earlier Models}

\citeauthor{arxiv:mosi} are among the first to advance this field, proposing TFN  \citep{emnlp:tfn}, which achieves comprehensive multimodal fusion through Cartesian products. As a variant of TFN, LMF  \citep{acl:lmf} is a more efficient model that uses a low-rank pattern. However, both methods neglect the temporal information of non-verbal modalities. Thus, they propose MFN  \citep{aaai:mfn} which addresses this issue by designing an LSTM  \citep{nc:lstm} system to capture temporal information. However, LSTM has multiple limitations in handling complex NLP tasks, particularly in representing long-range dependencies and complex temporal patterns, which has driven the development of Transformer \citep{nips:transformer}-based models that excel in these areas.

\subsection{Multimodal Fusion Oriented Models}

With the rise of Transformers, \citeauthor{acl:mult} proposed MulT \citep{acl:mult}, which, from the perspective of modality translation, effectively integrates multimodal data using Cross-Modal Attention (CMA) and Multi-Head Self Attention (MHSA) and implicitly aligns modality sequences. Building on MulT, CMA and MHSA, models such as TETFN  \citep{pr:tetfn}, ALMT  \citep{emnlp:almt}, and AcFormer  \citep{mm:acformer} focus on enhancing the representation capabilities of non-verbal modalities by leveraging the more comprehensive and stronger sentiment information contained in the text modality, thereby achieving superior representation and fusion capabilities. These models, categorized as MulTs, are among the most widely used and extensively validated approaches for multimodal fusion. As multimodal fusion is the core issue in MSA, MulTs are the backbones of a bunch of following works \citep{aaai:hhmpn, pr:tetfn, emnlp:almt, mm:acformer, naacl:mmml, ipm:cmhfm, eswa:fnenet, emnlp:mmim}.

\subsection{Finetuning Pretrained Transformers}

Except for the aforementioned, fine-tuning pre-trained Transformers (BERT\citep{naacl:bert}) with multimodal adaptation gates, such as in MAG-BERT \citep{acl:mag-bert} and its successors CENet \citep{tmm:cenet}, HyCon \citep{taffc:hycon}. 

\subsection{Representation Learning-based Models}

There are also models focusing on enhancing model robustness and representation through representation learning-based methods like MFM \citep{iclr:mfm}, Self-MM \citep{aaai:self-mm}, ConFEDE \citep{acl:confede}, and MTMD \citep{taffc:mtmd}, and combining multimodal Transformers with representation learning in models such as TETFN \citep{pr:tetfn} and MMML \citep{naacl:mmml}, have all shown significant improvements in MSA tasks

\subsection{Graph-based Models}

Graph-based models have gained significant attention in the MSA field. Representative approaches include pure graph neural network models such as GPFN \citep{acmtmm:gpfn}, which leverages graph convolution and pooling, and MTAG \citep{naacl:mtag}, which utilizes attention graphs. Additionally, graph theory-based Transformer models like HHMPN \citep{aaai:hhmpn}, a hierarchical model integrating MulT and message-passing routing, have also been explored.

Our proposed GsiT is a graph theory-based Transformer model. It combines the prior structural advantages of graph models for multimodal fusion with the powerful representational capacity of Transformers, effectively balancing efficiency and performance. Unlike traditional message-passing methods, our fusion process is executed in parallel, employing a prior structure designed as the Interlaced Mask.


\section{Weight Regularity}

\begin{figure*}[htpb]
\centering
  \includegraphics[width=\linewidth]{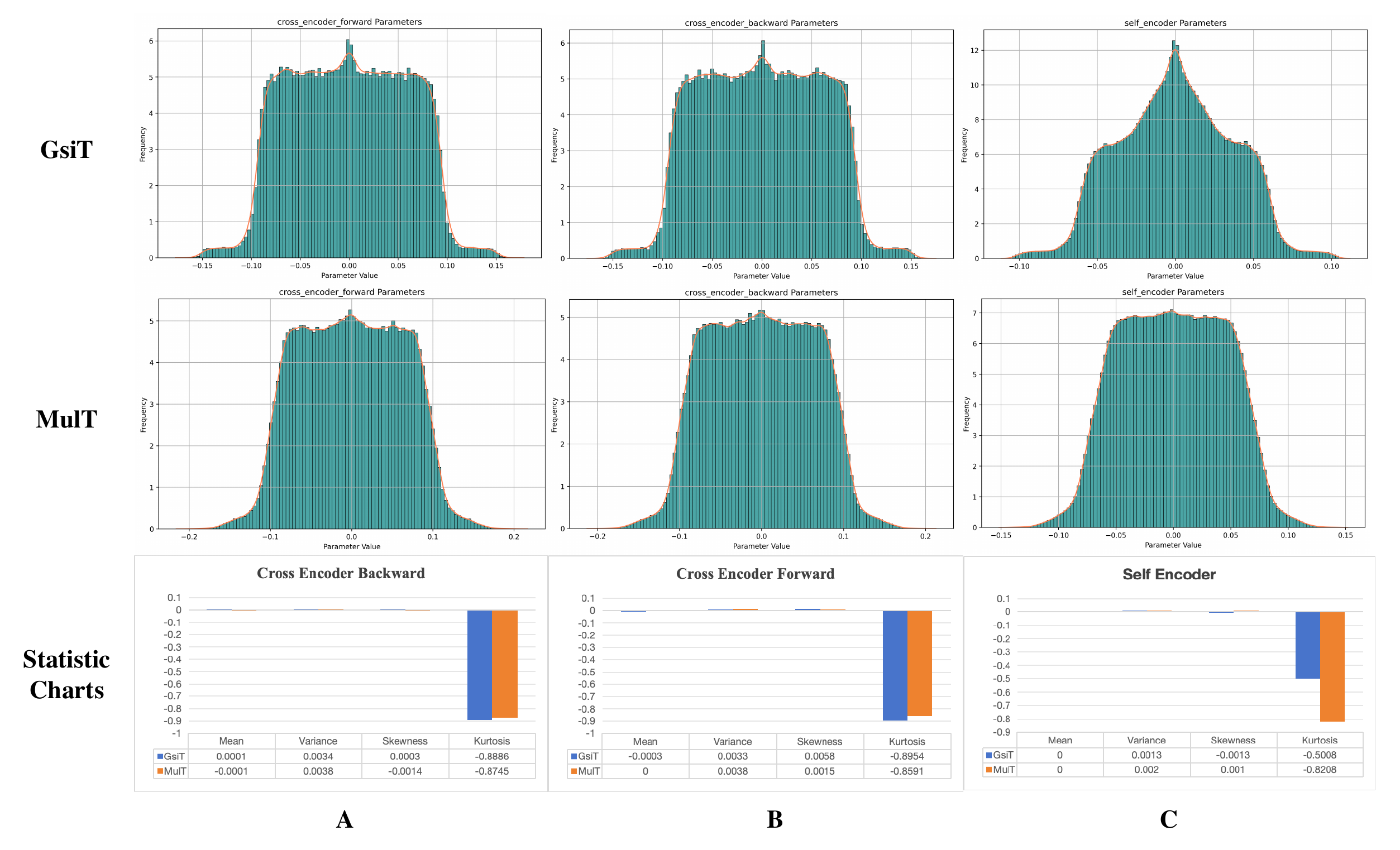}
  \caption{Parameter statistics of GsiT and MulT. A: Multimodal Fusion Encoder (backward); B: Multimodal Fusion Encoder (forward); C: Intra Enhancement Encoder.}
  \label{fig:weight regularity}
\end{figure*}

As shown in Figure \ref{fig:weight regularity}, both MulT and GsiT exhibit similar weight value distributions in the multimodal fusion encoders, with minimal differences, indicating a consistent multimodal fusion process in terms of weight distribution. However, in the intra-enhancement encoder, GsiT shows a notably lower kurtosis compared to MulT, suggesting that the weights are more evenly distributed and closer to a normal distribution. This indicates that GsiT has higher regularity, reducing the likelihood of overfitting and improving model generalization. To make the weight distributions comparable, we extracted corresponding combinations from MulT in a manner consistent with GsiT. Each combination set in MulT consists of three bi-modality combinations, and we analyzed the overall weight distribution of these sets.

\begin{figure}[htpb]
\centering
  \includegraphics[width=\linewidth]{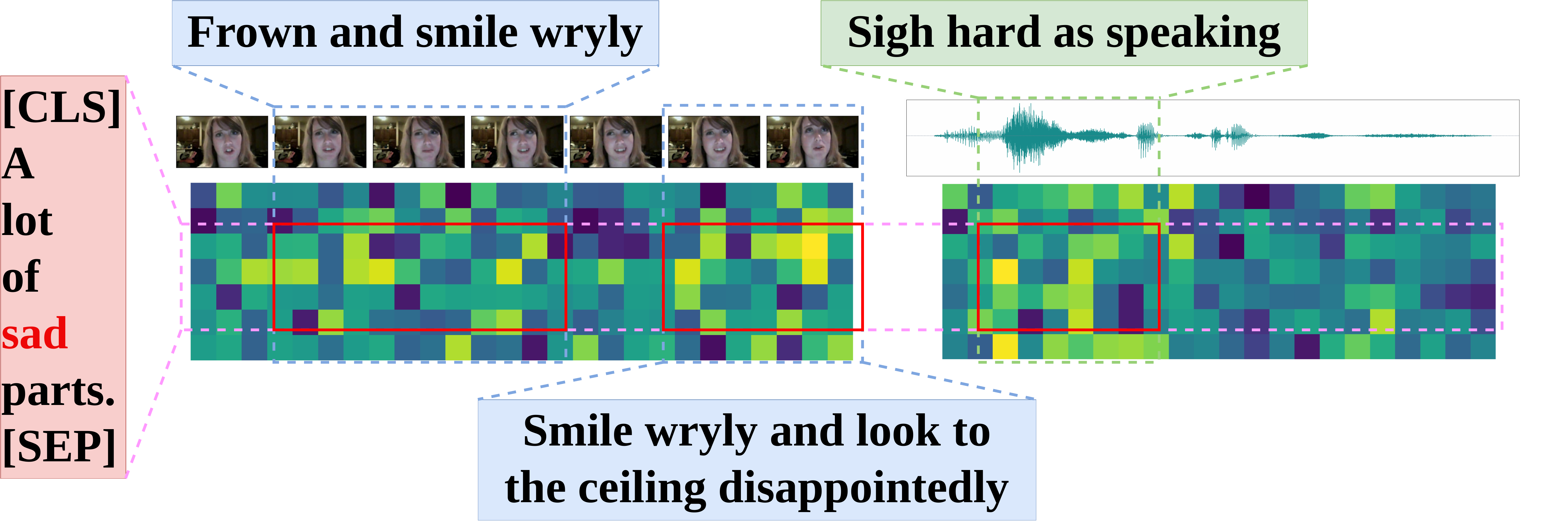}
  \caption{The realtime example of adjacency of GsiT.}
    \label{fig:exp of adjacency matrix}
\end{figure}

\begin{figure}[htpb]
\centering
  \includegraphics[width=\linewidth]{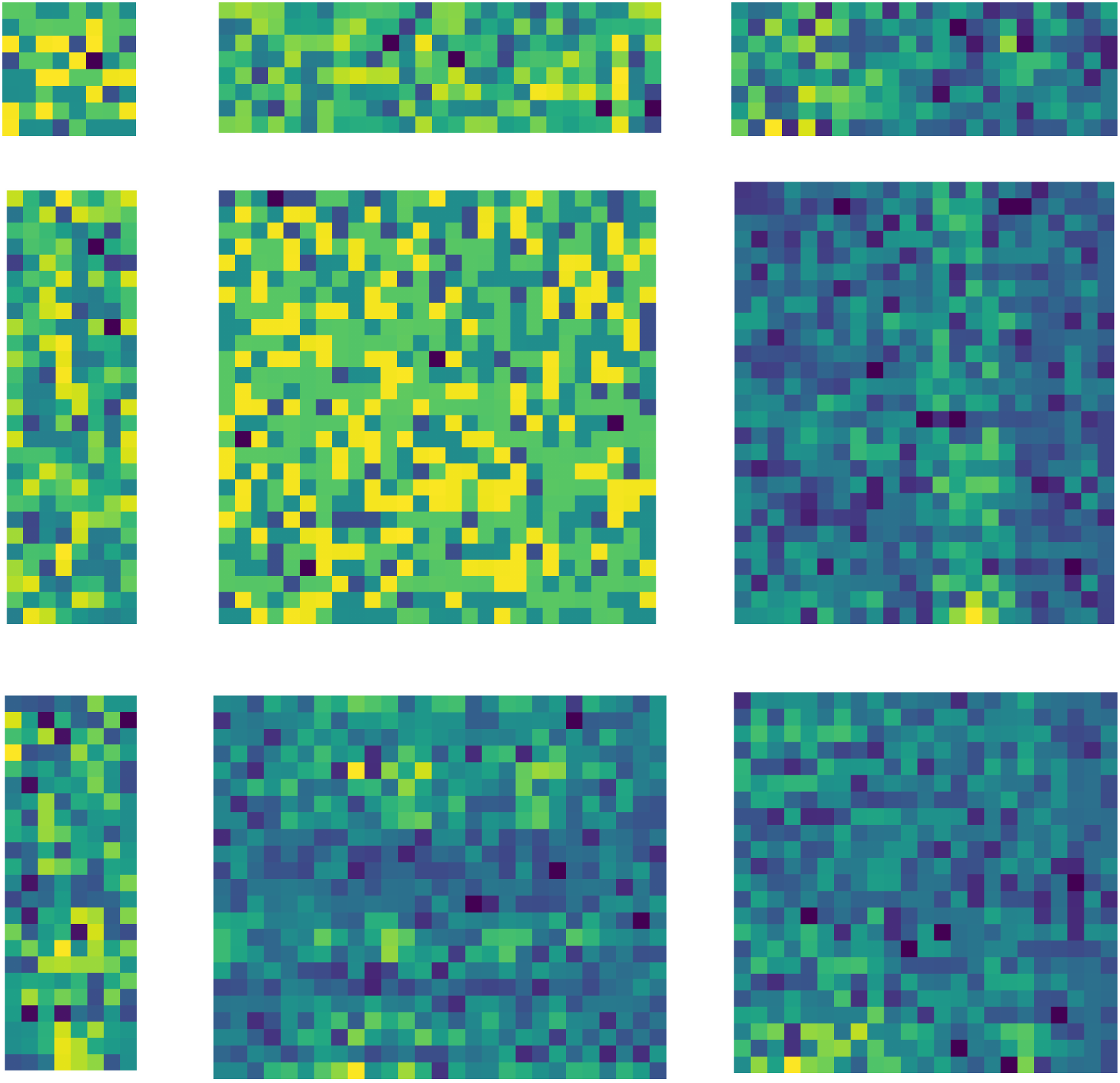}
  \caption{The adjacency matrix of GsiT.}
    \label{fig:adjacency matrix}
\end{figure}

\section{Adjacency Matrix}

Figure \ref{fig:exp of adjacency matrix} illustrates a real-time example of the adjacency structure in GsiT, showcasing its dynamic connectivity patterns. Figure \ref{fig:adjacency matrix} presents the adjacency matrix (attention map) of GsiT, visualizing the learned relationships and interactions within the model. Both figures highlight GsiT's ability to capture and represent complex dependencies effectively.

\end{document}